\pdfoutput=1

\documentclass[11pt]{article}

\usepackage[]{EMNLP2023}

\usepackage{times}
\usepackage{latexsym}

\usepackage[T1]{fontenc}

\usepackage[utf8]{inputenc}

\usepackage{microtype}

\usepackage{graphicx}
\usepackage{inconsolata}
\usepackage{multirow}
\usepackage{subfloat}
\usepackage{tabularray}
\usepackage{graphicx}
\usepackage{array}
\usepackage[caption=false,labelfont=sf,textfont=sf]{subfig}
\usepackage{textcomp}
\usepackage{amsfonts}
\usepackage{xcolor}

\UseTblrLibrary{booktabs}

\title{Parameter-Efficient Fine-Tuning with Layer Pruning on Free-Text Sequence-to-Sequence Modeling}

\author{
  Yunqi Zhu$^{1, 2}$ \enspace Xuebing Yang$^{2, *}$ \enspace Yuanyuan Wu$^{1}$ \enspace Wensheng Zhang$^{1, 2, 3, *}$ \\
  \small{$^{1}$School of Information and Communication Engineering, Hainan University} \\
  \small{$^{2}$State Key Laboratory of Multimodal Artificial Intelligence Systems, Institute of Automation, Chinese Academy of Sciences} \\
  \small{$^{3}$Guangzhou University} \\
  \small{\texttt{zhuyunqi96@163.com \,
                yangxuebing2013@ia.ac.cn \,
                wyuanyuan82@163.com \,
                zhangwenshengia@hotmail.com
}}}

\begin{document}
\maketitle

\renewcommand{\thefootnote}{\fnsymbol{footnote}}
\footnotetext[1]{Corresponding authors.}
\renewcommand{\thefootnote}{\arabic{footnote}}

\begin{abstract}
  The increasing size of language models raises great research interests in parameter-efficient fine-tuning such as LoRA that freezes the pre-trained model,
  and injects small-scale trainable parameters for multiple downstream tasks (e.g., summarization, question answering and translation).
  To further enhance the efficiency of fine-tuning,
  we propose a framework that integrates LoRA and structured layer pruning.
  The integrated framework is validated on two created deidentified medical report summarization datasets based on MIMIC-IV-Note and two public medical dialogue datasets.
  By tuning 0.6\% parameters of the original model and pruning over 30\% Transformer-layers,
  our framework can reduce 50\% of GPU memory usage and speed up 100\% of the training phase,
  while preserving over 92\% generation qualities on free-text sequence-to-sequence tasks \footnote{
    The source code and the dataset creation scripts are available at: \url{https://github.com/zhuyunqi96/LoraLPrun}
  }.
\end{abstract}

\section{Introduction}

Pre-trained language model (PLM) has dominated the natural language processing (NLP) domain because of its superior performance by fine-tuning the model to different downstream NLP tasks \cite{bert-model, gpt-1, bart-model, t5-model}.
With the increasing model size of PLMs (e.g., 11B for T5 \cite{t5-model}, 175B for GPT-3 \cite{gpt3-model} and 540B for PaLM \cite{palm-model}),
full fine-tuning of a PLM requires storing all the updated model parameters for each task,
which would be expensive and time-consuming.

\setlength{\belowcaptionskip}{-0.7em} 
\begin{figure}[t]
  \centering
  \includegraphics[width=1.0 \linewidth]{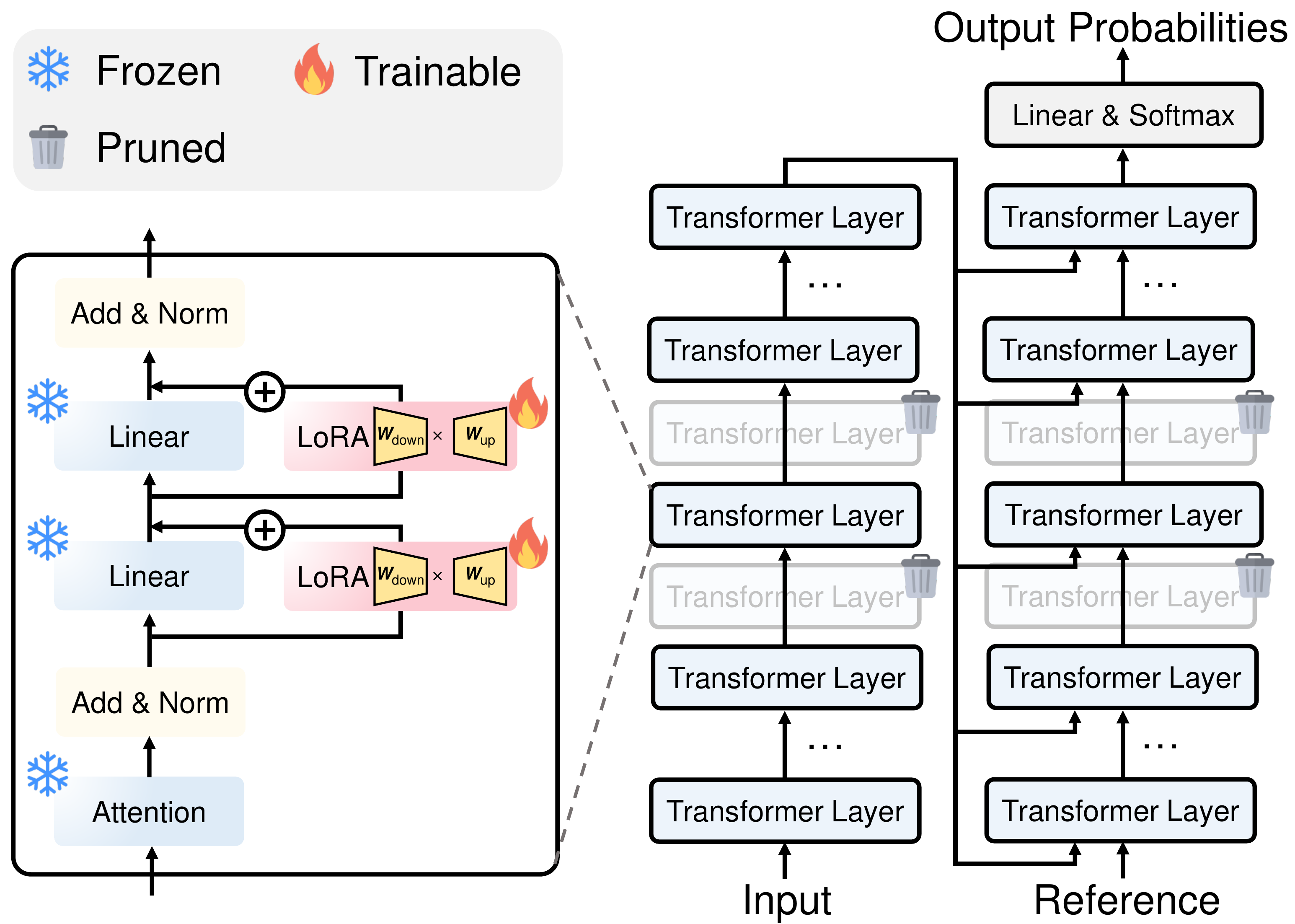}
  \vspace{-2.0em}
  \caption{
    Overview of the proposed framework.
    We freeze the pre-trained language model
    and remove the Transformer-layers in an alternating fashion,
    then LoRA is injected on the feed-forward networks.
  }
  \label{fig:model1}
\end{figure}
\begin{figure}[t]
  \centering
  \includegraphics[width=1.0 \linewidth]{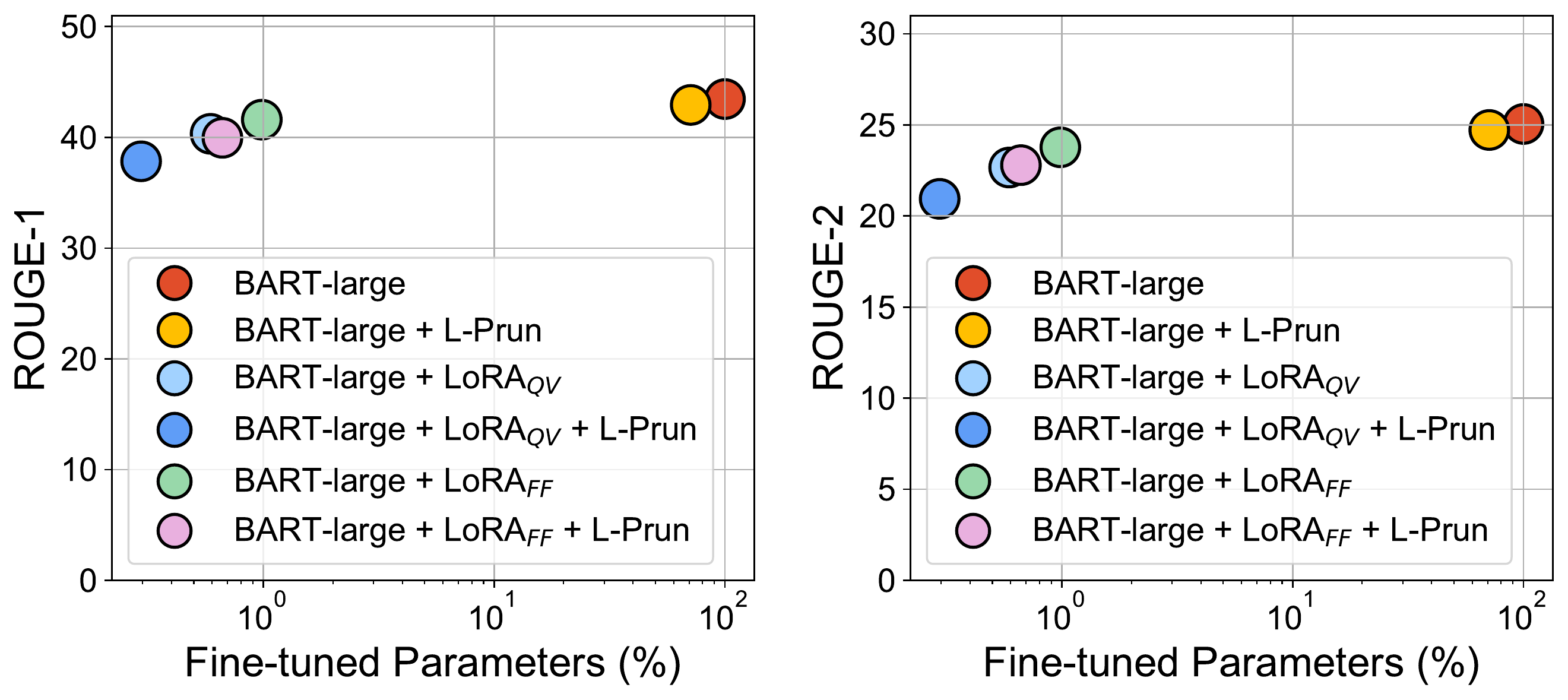}
  \vspace{-2.0em}
  \caption{
    ROUGE-1 and ROUGE-2 evaluations of full fine-tuning, LoRA and L-Prun with BART-large on MIMIC-IV-discharge.
  }
  \label{fig:dis-fig1}
\end{figure}

To alleviate the above issue, parameter-efficient fine-tuning approaches are emerging.
For instance, Adapter-based tuning \cite{adapter-model, mad-x-model, k-adapter-model, adapterfusion-model, compacter-model} injects small-scale trainable neural networks and freezes other parameters of a PLM during training.
Based on the prior knowledge of downstream tasks, prompt-based tuning \cite{prefix-tuning, p-tuning, prompt-tuning-largescale, p-tuning-v2} includes task-specific trainable contextual tokens around the input sequence and freezes the original input during training.
Recently, low-rank adaptation (LoRA) \cite{lora-model}, a variant based on Adapter-based tuning but removes nonlinear activation,
has become increasingly popular for small-scale parameter-tuning.
Besides, pruning of the PLM serves to remove some parameters of the model while largely perserving the capacity for downstream tasks \cite{tinybert, progressive-pruning, movement-pruning, layerdrop, layer-dropping, block-pruning, cofi-model},
which is usually processed in an adaptive or structured fashion.

Since both Adapter-based tuning and pruning hypothesize PLMs are over-parameterized,
we integrate LoRA \cite{lora-model} and structured layer pruning (L-Prun) \cite{layerdrop, layer-dropping} on PLMs,
and Figure~\ref{fig:model1} illustrates the overall framework of integrating LoRA and L-Prun on a Transformer-based \cite{Transformer-model} PLM.
Our proposal is to offer a preliminary framework that enables flexible and lightweight deployment of small-scale plug-in learned parameters with LoRA,
while scaling down PLMs with L-Prun.

The proposed framework is validated on two downstream tasks
(i.e., Medical Report Summarization and Medical Dialogue Generation).
Based on MIMIC-IV-Note \cite{mimic-iv-cite,mimic-iv-note-cite,mimic-iv-paper}, we firstly create two free-text medical deidentified report summarization datasets, named MIMIC-IV-discharge and MIMIC-IV-radiology.
Further, we implement extensive experiments with sequence-to-sequence (Seq2Seq) modeling, including BART-large \cite{bart-model} and T5-large \cite{t5-model}, on both two datasets and two extra publicly available medical dialogue datasets (HealthCareMagic and iCliniq) \cite{meddialog-paper, chatdoctor-paper}.
Figure~\ref{fig:dis-fig1} shows the performances of full fine-tuning, LoRA and L-Prun with pre-trained BART-large,
where the generation quality of our framework is promising.

Our contributions can be summarized as follows:
(1) We propose a framework that integrates small-scale parameter-tuning module LoRA and structured L-Prun.
To the best of our knowledge, we are the first to combine LoRA with L-Prun on PLMs.
(2) Based on MIMIC-IV-Note, we create two medical report summarization datasets.
We implement comprehensive experiments on the proposed datasets and the other two medical dialogue generation datasets.
(3) By updating 0.6\% parameters over the original PLM and pruning more than 30\% Transformer-layers,
the proposed method can double the training speed and save 50\% of GPU memory usage,
while perserving over 92\% generation qualities on medical Seq2Seq tasks.

\section{Related Work}

\paragraph{Transformer}
\cite{Transformer-model} proposed the Transformer model, containing an encoder and a decoder for the representation learning of input and output sequences, respectively.
The Transformer encoder layer consists of an attention block (ATTN), a feed-forward network (FFN), residual connections and layer normalizations.
The Transformer decoder layer contains an additional cross-attention block right after the original ATTN,
and the cross-attention has the same design as the self-attention.
Designed with up- and down-projection,
the FFN is two fully connected layers (i.e., multilayer perceptrons (MLPs)).

\paragraph{Parameter-Efficient Fine-Tuning}
Pre-training a large language model with self-supervised language modeling and fine-tuning it to multiple downstream domains is a widespread paradigm \cite{bert-model, gpt-1, bart-model, t5-model}.
\cite{delta-tuning} categorize popular parameter-efficient fine-tuning methods into three types:
\textit{Addition}, \textit{Specification} and \textit{Reparameterization}.

\textit{Addition}: Adapter-based tuning \cite{adapter-model, mad-x-model, k-adapter-model, adapterfusion-model, compacter-model}
injects lightweight down- and up-projection MLPs and a nonlinear activation function,
and prompt-based tuning \cite{prefix-tuning, p-tuning, prompt-tuning-largescale, p-tuning-v2, pretrain-prompt-tuning} freezes the input and introduces additional trainable contexts (based on the prior knowledge of NLP tasks) as the prefix, suffix or semi-template for the original input.

\textit{Specification}: fine-tuning only specified or adaptive limited number of layers, modules, weights and biases of the model \cite{elsa-freezing, masking-efficient, diff-pruning, mae-linear-probing, bitfit},
this type of methods can be the most memory-efficient for training because no additional parameter is involved.

\textit{Reparameterization}: based on the hypothesis that adaptations of the PLM to downstream NLP tasks can be reparameterized into a low intrinsic rank optimization \cite{residual-adapters, intrinsic-dim, intrinsic-low-rank},
LoRA \cite{lora-model} inherits the design of Adapter \cite{adapter-model} but removes the nonlinear activation.
Therefore, the multiplied weights of down- and up-projection MLPs can be directly added to the weight of the injection target in the inference phase.

In addition, there are alternative methods to efficiently leverage the pre-training language model,
including Knowledge Distillation \cite{distill-model, seq-knowledge-distill, distill-bert, tinybert} that maximizes the similarity between the teacher (large model) and the student's (distilled model) predictions,
Quantization \cite{quantization-2018, quantization-training, quantization-llm-8bit} that converts a 16/32-bit model to an 8-bit or even lower one,
which is both efficient for training and inferencing.

\paragraph{Parameter Pruning}
A large pre-trained model can be over-parameterized for downstream NLP tasks,
\cite{tinybert, progressive-pruning, movement-pruning, layerdrop, layer-dropping, block-pruning, cofi-model} showed that structured and adaptive pruning of the PLM can be competent enough for many classification or Seq2Seq scenarios.
Structured pruning can be based on an empirical or heuristic choice of the model,
whilst unstructured pruning may rely on a few pre-defined thresholds or trainable parameters to adaptively decide which MLP, attention block or the entire layer should be dropped.

Additionally, alternative parameter pruning methods like dropping or merging the hidden state of unimportant tokens of the sequence \cite{power-bert, longt5, learned-token-pruning, token-dropping}, and implementing sparse attention \cite{sparse-Transformer, longformer, reformer} contribute to decrease memory usage but could not decrease the model size.

\section{Methodology}

We propose a hybrid framework of LoRA and structured layer pruning that can considerably decrease the trained parameters,
lower the memory usage, increase the training speed,
and compress the model size for downstream tasks.
The motivation for this integration is to provide a preliminary framework that can scale down the PLM on downstream NLP tasks through L-Prun,
and enable flexible and lightweight deployment of small-scale plug-in trained parameters through LoRA.
We first delete the Transformer-layers in an alternating fashion and then inject LoRA for the model,
hence the method does not require a whole adaptive pruning stage after the model is fine-tuned.
Visualization of the framework is shown in Figure~\ref{fig:model1}.
Formally, given a PLM with $P_{m}$ parameters,
LoRA enables $P_{m: \ lora}$ trainable parameters, where $P_{m: \ lora} \ll P_{m}$,
and L-Prun makes the parameters of PLM shrink to $P_{prun}$, where $P_{prun} < P_{m}$.
For $N$ downstream tasks, the overall storage is $N \times P_{m}$ for full fine-tuning,
$N \times P_{m: \ lora} + P_{m}$ if LoRA is enabled,
and $N \times P_{prun: \ lora} + P_{prun}$ if both LoRA and L-Prun are enabled.
Note that $P_{prun: \ lora} < P_{m: \ lora}$ due to LoRA is injected into all Transformer-layers.
In the following, we briefly introduce LoRA and Layer Pruning, as well as clarify our implementation.

\paragraph{LoRA}
\cite{lora-model} proposed a low-rank adaptation method that freezes the weights of a Transformer-based PLM,
and injects two trainable dense layers ($W_{down} \in \mathbb{R}^{r \times d}$ and $W_{up} \in \mathbb{R}^{d \times r}$) in a dense layer $W_{0} \in \mathbb{R}^{d \times d}$,
where the rank $r \ll d$.
Therefore the output $\widetilde{x}$ of LoRA is:
\begin{equation}
  \widetilde{x} = W_{0}x + \frac{\alpha}{r} W_{up}W_{down}x
  \label{eq:cross-attn}
\end{equation}
where $\alpha$ is a constant scaling factor.
LoRA can be viewed as a variant of Adapter \cite{adapter-model} containing down- and up-projection MLPs.
However, unlike conventional Adapter methods applying the nonlinear activation between two trainable dense layers,
LoRA can retain a low inference latency after the training phase by merging the original frozen weights with LoRA's injection:
\begin{equation}
  \widetilde{W_{0}} = W_{0} + \frac{\alpha}{r} W_{up}W_{down}
\end{equation}
With the down- and up-projection, LoRA has much less parameters than the injection target.

Originally, LoRA (LoRA$_{QV}$) was applied on the Query dense layer and the Value dense layer in each of the attention module.
Further researches \cite{adalora-model, conditional-adpater} showed that injecting LoRA (LoRA$_{FF}$) in FFN could be a superior choice.
Recently, Adapter-based parameter pruning researches \cite{adapterdrop-model, splora, adalora-model, prune-lora} focus on adaptively removing redundant trainable parameters within the LoRA unit,
and adaptively skipping unimportant tokens' hidden states \cite{conditional-adpater},
and exploring the optimal configuration of parameter-efficient learning methods \cite{adamix, uni-peft-model}.

\paragraph{Layer Pruning}
\cite{layerdrop, layer-dropping} proposed strategies of structured pruning of Transformer-layers inside a Transformer-based PLM for fine-tuning downstream tasks.
The empirical experiments show that deleting the Transformer-layers in an alternating fashion and preserving a few top layers and bottom layers can sustain the capacity of language models.
Meanwhile, due to the reduction of the model size,
the training phase and the inference phase can be greatly accelerated.
We symmetrically drop the Transformer-layers in the encoder and the decoder,
for a PLM with $n$ layers of the encoder and $n$ layers of the decoder:
$\{ l_{1}, ... , l_{n} \}$.
Denotes the dropped layers as $l_{i}$, where $ i = 2k, k \in \mathbb{N}, i \in [4, n-2]$.
Therefore, a total of $ 2(n - 4) $ layers are dropped.

\section{Experiments}

\subsection{Datasets}

\begin{table*}
  \centering
  \scalebox{0.98}{
    \begin{tblr}{colspec={lccccccc},
      cell{1}{1}={r=2}{}, cell{1}{2}={r=2}{}, cell{1}{3,5,7}={c=2}{},
          rowsep=0.5pt, stretch=1, rows={ht=\baselineskip}}
      \toprule
      Dataset            & Train/Eval/Test      & Avg Input &               & Avg Output &         & \% novel &        \\
      \cmidrule[lr]{3-4} \cmidrule[lr]{5-6} \cmidrule[lr]{7-8}
                         &                      & Sents     & Words         & Sents      & Words   & unigram  & bigram \\
      \hline
      MIMIC-IV-discharge & 254K / 31K / 31K     & 88.5      & 1879.9        & 1.1        & \, 13.8 & 18       & 58     \\
      MIMIC-IV-radiology & 1.2M / 0.15M / 0.15M & 10.0      & \, 132.1      & 2.9        & \, 38.4 & 40       & 73     \\
      HealthCareMagic    & 165K / 20K / 20K     & \, 4.9    & \,\,\,\, 89.9 & 6.6        & \, 90.7 & 80       & 98     \\
      iCliniq            & 12k / 1.5K / 1.5K    & \, 7.2    & \, 111.3      & 9.4        & 126.8   & 77       & 97     \\
      \bottomrule
    \end{tblr}
  }
  \vspace{-0.5em}
  \caption{Statistics of the datasets. \% novel N-gram represents the ratio of unseen N-grams from the reference output against the source input.}
  \label{tab:dataset-stats}
\end{table*}

Based on MIMIC-IV-Note \cite{mimic-iv-cite,mimic-iv-note-cite,mimic-iv-paper},
a publicly available deidentified free-form clinical notes with 331K discharge reports and 2.3M radiography reports at the Beth Israel Deaconess Medical Center in Boston, MA, USA,
we create two medical report summarization datasets with regular string pattern matching over the report subsection titles,
named MIMIC-IV-discharge and MIMIC-IV-radiology.

We conduct extensive experiments on four medical datasets,
including MIMIC-IV-discharge, MIMIC-IV-radiology, HealthCareMagic \cite{meddialog-paper, chatdoctor-paper} and iCliniq \cite{meddialog-paper, chatdoctor-paper}.
Statistics of the four datasets are shown in Table~\ref{tab:dataset-stats}.

\paragraph{Summarization}
We use MIMIC-IV-discharge and MIMIC-IV-radiology for medical report summarization.
\textit{(i)} MIMIC-IV-discharge:
from MIMIC-IV-Note,
a discharge report would involve the notes of addmision, medical history, health care, procedure, etc.
The corresponding discharge summary contains the overall diagnosis of the patient.
\textit{(ii)} MIMIC-IV-radiology:
from MIMIC-IV-Note,
written with semi-structured templates, a radiography report contains the full description of the medical imaging results,
while the corresponding summary includes free-text highlights of the report.

\paragraph{Dialogue}
We use HealthCareMagic and iCliniq for medical dialogue generation.
\textit{(i)} HealthCareMagic: a patient-doctor conversations dataset crawled from a online medical consultation platform \footnote{\url{healthcaremagic.com}}.
We used a ``HealthCareMagic-200K'' version from \cite{chatdoctor-paper}.
\textit{(ii)} iCliniq: a patient-doctor conversations dataset crawled from an online medical consultation platform \footnote{\url{iclinic.com}}.
We used a ``iCliniq-15K'' version \cite{chatdoctor-paper}.

\definecolor{c_grey1}{RGB}{235, 235, 235}
\begin{table}
  \centering
  \scalebox{0.38}{
    \begin{tblr}{colspec={ *{1}{t{19mm}}*{1}{m{12mm}}*{24}{m{2.5mm}} },
      cell{1}{1}={r=2}{}, cell{3}{1}={r=2}{},
      cell{1,2}{3,4,5,7,9,11,13,14} = {c_grey1},
      cell{3,4}{3,4,5,7,9,11,13,15,17,19,21,23,25,26} = {c_grey1},
          stretch=1, rows={ht=\baselineskip}}
      \toprule
      BART-large & Encoder & 1 & 2 & 3 & 4 & 5 & 6 & 7 & 8 & 9 & 10 & 11 & 12 &    &    &    &    &    &    &    &    &    &    &    &    \\
                 & Decoder & 1 & 2 & 3 & 4 & 5 & 6 & 7 & 8 & 9 & 10 & 11 & 12 &    &    &    &    &    &    &    &    &    &    &    &    \\
      \hline
      T5-large   & Encoder & 1 & 2 & 3 & 4 & 5 & 6 & 7 & 8 & 9 & 10 & 11 & 12 & 13 & 14 & 15 & 16 & 17 & 18 & 19 & 20 & 21 & 22 & 23 & 24 \\
                 & Decoder & 1 & 2 & 3 & 4 & 5 & 6 & 7 & 8 & 9 & 10 & 11 & 12 & 13 & 14 & 15 & 16 & 17 & 18 & 19 & 20 & 21 & 22 & 23 & 24 \\
      \bottomrule
    \end{tblr}
  }
  \vspace{-0.5em}
  \caption{Remaining Transformer-layers of BART-large $+$ L-Prun and T5-large $+$ L-Prun are highlighted in grey.}
  \label{tab:l-prun-pattern}
\end{table}

\subsection{Implementation Details}

We implement the experiments with the pre-trained BART-large and T5-large.
We set hyperparameter rank $r$ as 16 and factor $\alpha$ as 32 for LoRA in this work.
For $r$, additional experiments (Appendix~\ref{sec:appendix-hyper} Table~\ref{tab:hyper-alpha}) show that 16 will be a balanced and competent choice.
In Table~\ref{tab:l-prun-pattern}, we show the specific perserving indexes of encoder and decoder layers for BART-large and T5-large.
Hence, 33\% and 41\% of the Transformer-layers are dropped for BART-large and T5-large.
Next, based on the empirical hyperparameter settings,
we use PyTorch framework with mixed-precision,
AdamW \cite{adamW} optimizer ($\beta_{1}$ = 0.1, $\beta_{2}$ = 0.999, $\epsilon$ = $10^{-8}$)
and warm-up steps of 1000.
We fine-tune the original model and L-Prun model at a learning rate of $5 \times 10^{-5}$,
and fine-tune the model with LoRA at a learning rate of $1 \times 10^{-4}$.
Furthermore, the maximum source lengths of 1024 and 512 are set for summarization and dialogue datasets, respectively.
All of the maximum target lengths are 128.
A batch size of 8 is applied for all the experiments.
However, we use a setting: (batch size: 4; gradient accumulation: 2) only when fine-tuning T5-large on MIMIC-IV-discharge because of limited GPU memory,
while the gradient accumulation is disabled in other cases.
We fine-tune the datasets with 10 epochs.
For every 0.3 epoch, the model is evaluated, and the checkpoint with the highest ROUGE-1 score is loaded for the test set.
We use beam search during the autoregressive decoding with a beam width of 6.
All the experiments are conducted on a single NVIDIA A40 48GB GPU.

\subsection{Evaluation Metrics}

The machine-generated outputs are evaluated with the following metrics.
Note that the machine-generated summaries are evaluatd with ROUGE, BERTScore and SummaC,
while the machine-generated dialogues are evaluatd with ROUGE, BERTScore and BLEU.

\paragraph{ROUGE}
\cite{rouge-metric} is an N-gram-based recall-oriented metric popular for evaluating automatic summarization.
R-1, R-2 and R-L represent the overlapping degree of unigram, bigram and longest common subsequence between the candidate and the reference, respectively.

\paragraph{BERTScore}
\cite{bertscore-metric} is a contextual semantic evaluation metric based on a pre-trained BERT,
which compares the cosine similarity between the latent representation of the \texttt{[CLS]} token of a candidate with a reference's.

\paragraph{SummaC}
\cite{summac-metric} is a reference-free lightweight factual consistency evaluation metric for automatic summarization.
Using fine-tuned natural language inference (NLI) model, the framework detects the sentence-level inconsistency between the source document and the machine-generated summary.

\paragraph{BLEU}
\cite{bleu-metric} is a popular N-gram-based precision-oriented evaluation metric that compares the overlapping degree in tokens between the machine-generated text and reference text.
BLEU-$N$ indicates the weighted average of the N-grams matching evaluation, $N = \{1,...,N\}$,
and we report BLEU-1 and BLEU-4 for the dialogue datasets.

\subsection{Baselines and Parameters}

We implement the proposed method with BART-large and T5-large:

\paragraph{BART-large}
\cite{bart-model}: a Transformer-based encoder-decoder model pre-trained on a combination of English books, news, stories and Wikipedia paragraphs with masked language modeling.
BART-large (406M parameters) has 12 layers of encoder, 12 layers of decoder, a hidden state size of 1024 and a vocabulary size of 50K.

\paragraph{T5-large}
\cite{t5-model}: a Transformer-based encoder-decoder model pre-trained on a large clean corpus (C4) in English with masked language modeling,
the pre-trained model implements multiple NLP training tasks info a text-to-text paradigm.
T5-large (770M parameters) contains 24 layers of encoder, 24 layers of decoder, a hidden state size of 1024 and a vocabulary size of 32K.

\subsection{Experimental Results}

\begin{table}[t]
  \centering
  \scalebox{0.6}{
    \begin{tblr}{colspec={lcrrcc}, rowsep=0.5pt, stretch=1.1, rows={ht=\baselineskip}, row{5, 7} = {bg=c_grey1}}
      \toprule
      Model                      & Speed  & Trained Params & Mem Used & R-1   & R-2   \\
      \hline
      BART-large                 & 100 \% & 406.2 M        & 34.56 GB & 43.46 & 25.05 \\
      $+$ L-Prun                 & 145 \% & 288.7 M        & 23.65 GB & 42.93 & 24.72 \\
      $+$ LoRA$_{QV}$            & 115 \% & 2.4 M          & 28.15 GB & 40.32 & 22.66 \\
      $+$ LoRA$_{QV}$ $+$ L-Prun & 170 \% & 1.2 M          & 19.13 GB & 37.84 & 20.94 \\
      $+$ LoRA$_{FF}$            & 125 \% & 4.0 M          & 27.25 GB & 41.60 & 23.77 \\
      $+$ LoRA$_{FF}$ $+$ L-Prun & 179 \% & 2.7 M          & 17.99 GB & 39.96 & 22.79 \\
      \bottomrule
    \end{tblr}
  }
  \vspace{-0.5em}
  \caption{Comparison of different LoRA insertion methods and L-Prun methods in BART-large on MIMIC-IV-discharge.}
  \label{tab:method-comparison-dis}
\end{table}

\begin{table*}[!htbp]
  \subfloat[\small{MIMIC-IV-discharge}\label{result-dis}]{\scalebox{0.875}{
      \begin{tblr}{colspec={lcrrccccc}, rowsep=0.5pt, stretch=1.0, rows={ht=\baselineskip}, row{5, 9} = {bg=c_grey1}}
        \toprule
        Model                      & Speed  & Trained Params & Mem Used & R-1   & R-2   & R-L   & BERTScore & SummaC \\
        \hline
        BART-large                 & 100 \% & 406.2 M        & 34.56 GB & 43.46 & 25.05 & 43.03 & 87.13     & 56.45  \\
        $+$ L-Prun                 & 145 \% & 288.7 M        & 23.65 GB & 42.93 & 24.72 & 42.48 & 87.09     & 56.70  \\
        $+$ LoRA$_{FF}$            & 125 \% & 4.0 M          & 27.25 GB & 41.60 & 23.77 & 41.20 & 86.85     & 56.25  \\
        $+$ LoRA$_{FF}$ $+$ L-Prun & 179 \% & 2.7 M          & 17.99 GB & 39.96 & 22.79 & 39.54 & 86.44     & 56.65  \\
        \hline
        T5-large                   & 100 \% & 737.6 M        & 42.49 GB & 41.22 & 24.09 & 38.82 & 85.80     & 55.24  \\
        $+$ L-Prun                 & 169 \% & 443.9 M        & 25.12 GB & 40.70 & 23.94 & 38.46 & 85.71     & 55.86  \\
        $+$ LoRA$_{FF}$            & 115 \% & 8.1 M          & 31.15 GB & 38.75 & 22.34 & 36.79 & 85.26     & 57.97  \\
        $+$ LoRA$_{FF}$ $+$ L-Prun & 197 \% & 4.7 M          & 17.97 GB & 37.86 & 21.79 & 35.99 & 85.03     & 56.28  \\
        \bottomrule
      \end{tblr}}}
  \vspace{-0.6em}
  \subfloat[\small{MIMIC-IV-radiology}\label{result-rad}]{\scalebox{0.875}{
      \begin{tblr}{colspec={lcrrccccc}, rowsep=0.5pt, stretch=1.0, rows={ht=\baselineskip}, row{5, 9} = {bg=c_grey1}}
        \toprule
        Model                      & Speed  & Trained Params & Mem Used & R-1   & R-2   & R-L   & BERTScore & SummaC \\
        \hline
        BART-large                 & 100 \% & 406.2 M        & 27.01 GB & 58.35 & 41.08 & 57.06 & 91.62     & 38.83  \\
        $+$ L-Prun                 & 142 \% & 288.7 M        & 18.55 GB & 58.34 & 41.09 & 57.05 & 91.63     & 39.04  \\
        $+$ LoRA$_{FF}$            & 136 \% & 4.0 M          & 20.24 GB & 55.20 & 37.40 & 53.91 & 91.06     & 40.75  \\
        $+$ LoRA$_{FF}$ $+$ L-Prun & 202 \% & 2.7 M          & 13.43 GB & 54.02 & 36.36 & 52.79 & 90.89     & 40.88  \\
        \hline
        T5-large                   & 100 \% & 737.6 M        & 30.98 GB & 57.19 & 40.05 & 55.00 & 89.56     & 42.88  \\
        $+$ L-Prun                 & 162 \% & 443.9 M        & 18.46 GB & 56.53 & 39.34 & 54.36 & 89.48     & 42.73  \\
        $+$ LoRA$_{FF}$            & 153 \% & 8.1 M          & 19.87 GB & 53.37 & 35.39 & 51.12 & 88.99     & 44.29  \\
        $+$ LoRA$_{FF}$ $+$ L-Prun & 206 \% & 4.7 M          & 11.66 GB & 52.22 & 34.36 & 50.03 & 88.90     & 44.33  \\
        \bottomrule
      \end{tblr}}}
  \vspace{-0.6em}
  \caption{
    Experiments on medical report summarization datasets.
  }
  \label{tab:result-table-mimic-iv}
\end{table*}
\begin{table*}
  \subfloat[\small{HealthCareMagic}\label{result-hcm}]{\scalebox{0.875}{
      \begin{tblr}{colspec={lcrrccccc}, rowsep=0.5pt, stretch=1.0, rows={ht=\baselineskip}, row{5, 9} = {bg=c_grey1}}
        \toprule
        Model                      & Speed  & Trained Params & Mem Used & R-1   & R-2  & R-L   & BLEU-1 & BLEU-4 \\
        \hline
        BART-large                 & 100 \% & 406.2 M        & 15.09 GB & 27.43 & 7.16 & 25.30 & 27.57  & 3.03   \\
        $+$ L-Prun                 & 127 \% & 288.7 M        & 10.59 GB & 27.38 & 7.16 & 25.28 & 27.78  & 3.03   \\
        $+$ LoRA$_{FF}$            & 162 \% & 4.0 M          & 9.47 GB  & 26.75 & 6.48 & 24.75 & 27.38  & 2.42   \\
        $+$ LoRA$_{FF}$ $+$ L-Prun & 211 \% & 2.7 M          & 6.51 GB  & 26.70 & 6.56 & 24.79 & 27.80  & 2.43   \\
        \hline
        T5-large                   & 100 \% & 737.6 M        & 31.01 GB & 26.97 & 6.83 & 24.84 & 26.92  & 2.67   \\
        $+$ L-Prun                 & 158 \% & 443.9 M        & 18.48 GB & 26.98 & 6.82 & 24.93 & 26.82  & 2.65   \\
        $+$ LoRA$_{FF}$            & 132 \% & 8.1 M          & 19.87 GB & 27.68 & 6.40 & 25.74 & 27.71  & 1.56   \\
        $+$ LoRA$_{FF}$ $+$ L-Prun & 222 \% & 4.7 M          & 11.66 GB & 27.27 & 6.41 & 25.56 & 29.04  & 1.81   \\
        \bottomrule
      \end{tblr}}}
  \vspace{-0.6em}
  \subfloat[\small{iCliniq}\label{result-icliniq}]{\scalebox{0.875}{
      \begin{tblr}{colspec={lcrrccccc}, rowsep=0.5pt, stretch=1.0, rows={ht=\baselineskip}, row{5, 9} = {bg=c_grey1}}
        \toprule
        Model                      & Speed  & Trained Params & Mem Used & R-1   & R-2  & R-L   & BLEU-1 & BLEU-4 \\
        \hline
        BART-large                 & 100 \% & 406.2 M        & 17.41 GB & 28.19 & 7.66 & 26.35 & 34.71  & 2.91   \\
        $+$ L-Prun                 & 138 \% & 288.7 M        & 12.21 GB & 28.13 & 7.32 & 26.32 & 33.57  & 2.79   \\
        $+$ LoRA$_{FF}$            & 158 \% & 4.0 M          & 11.65 GB & 26.07 & 6.49 & 24.56 & 33.18  & 2.73   \\
        $+$ LoRA$_{FF}$ $+$ L-Prun & 230 \% & 2.7 M          & 7.98 GB  & 27.35 & 6.54 & 25.81 & 32.25  & 2.25   \\
        \hline
        T5-large                   & 100 \% & 737.6 M        & 35.19 GB & 27.63 & 7.39 & 25.96 & 34.98  & 3.23   \\
        $+$ L-Prun                 & 164 \% & 443.9 M        & 20.92 GB & 28.58 & 7.47 & 26.86 & 33.87  & 2.96   \\
        $+$ LoRA$_{FF}$            & 128 \% & 8.1 M          & 23.96 GB & 28.25 & 6.77 & 26.43 & 32.94  & 2.25   \\
        $+$ LoRA$_{FF}$ $+$ L-Prun & 217 \% & 4.7 M          & 13.98 GB & 27.65 & 6.82 & 25.93 & 34.50  & 2.26   \\
        \bottomrule
      \end{tblr}}}
  \vspace{-0.6em}
  \caption{
    Experiments on medical dialogue datasets.
  }
  \label{tab:result-table-med-dialogue}
\end{table*}

\paragraph{LoRA$_{QV}$ or LoRA$_{FF}$}
In Table~\ref{tab:method-comparison-dis}, we show the performances of inserting LoRA in Query and Key verses FFN (i.e., LoRA$_{QV}$ and LoRA$_{FF}$).
``Speed'' refers to the relative speed of the training phase,
and the original full fine-tuning is considered as the speed baseline.
Since a Transformer decoder layer has two attention blocks and one FFN, LoRA$_{QV}$ is rather slower than LoRA$_{FF}$ in the training phase.
BART-large $+$ LoRA$_{FF}$ $+$ L-Prun is the overall practical solution for fast fine-tuning, low GPU memory usage, and good summarization quality.
Therefore, the following experimental results correspond to LoRA$_{FF}$ with L-Prun.
Note that the metrics Speed, R1, R2, R-L, BERTScore, Summac are the higher the better.

Next, we present the experimental result in Table~\ref{result-dis} (MIMIC-IV-discharge),
Table~\ref{result-rad} (MIMIC-IV-radiology),
Table~\ref{result-hcm} (HealthCareMagic) and
Table~\ref{result-icliniq} (iCliniq).

\paragraph{Medical Report Summarization}
In Table~\ref{result-dis} and Table~\ref{result-rad},
the evaluations among ROUGE and BERTScore show that applying L-Prun alone does not significantly drop the performance of either BART-large or T5-large.
Second, employing LoRA and L-Prun merely consume 68\% and 75\% as much GPU memory as the original full fine-tuning, respectively.
Further, the result of SummaC is getting better when the overall ROUGE and BERTScore are decreasing,
however it could be because the factual consistency metric trained on general corpus cannot meet the requirement of the medical domain.
Furthermore, the integration of LoRA and L-Prun enables the models to train 0.6\% parameters over the original model and to decrease 33\% or 41\% Transformer-layers of BART-large or T5-large,
while only a roughly 8\% performance drop with ROUGE scores and BERTScore is observed compared to the original full fine-tuning.

\paragraph{Medical Dialogue Generation}
As shown in Table~\ref{result-hcm} and Table~\ref{result-icliniq},
the dialogue generation datasets is more challenging than the summarization's because the \% novel N-gram of the reference output for the dialogue datasets is almost 100\% (Table~\ref{tab:dataset-stats}),
but the proposed method only brings marginal performance degradation (< 5\%) on R-1 and BLEU-1.
In addition, the overall speedup and GPU memory saving are consistent with the summarization's,
and up to $+$ 122\% faster training speed is accomplished on HealthCareMagic.
\paragraph{}
Overall, the key results show that the proposed method is a promising energy-saving approach.
Take BART-large as an example,
the total saved parameters for four Seq2Seq tasks with different methods are shown as follows:

\begin{itemize}
  \setlength{\parskip}{-0.4em}
  \item Full fine-tuning: \hspace{2.316em} 4 $\times$ 406.2M
  \item $+$ LoRA$_{FF}$ : \hspace{3.5448em} 4 $\times$ 4.0M $+$ 406.2M
  \item $+$ L-Prun: \hspace{4.71em} 4 $\times$ 288.7M
  \item $+$ LoRA$_{FF}$ $+$ L-Prun: 4 $\times$ 2.7M $+$ 288.7M
\end{itemize}


\section{Conclusion}
We propose a parameter-efficient fine-tuning framework that integrates LoRA and L-Prun.
Based on the assumption PLMs are over-parameterized for many downstream NLP tasks,
the proposed method uses small-scale trainable parameters to leverage PLMs,
and reduce the overall model size in a structured layer pruning way.
Further, we created two medical report summarization datasets from MIMIC-IV-Note,
and validated the proposed method on medical report summarization and medical dialogue generation.
By fine-tuning 0.6\% parameters of the PLM , dropping more than 30\% Transformer-layers in the PLM,
PLMs can sustain over 92\% generation qualities on medical Seq2Seq tasks and double the speed of training.

\section*{Limitations}

\textit{(i)} This study presents the integration of LoRA and structured L-Prun,
an automated self-adaptive structured L-Prun or block-pruning framework with Adapter/LoRA-based parameter-efficient fine-tuning would be worthy of further research.
\textit{(ii)} The proposed framework can be orthogonal to the other efficient fine-tuning approaches, such as Knowledge Distillation and Quantization.
Exploring an holistic integration paradigm for better energy-saving training is critical for future studies.
\textit{(iii)} This study implements the proposed framework on medical text-to-text datasets,
further research could focus on the parameter-efficient fine-tuning of multi-modal generation tasks.

\section*{Ethics Statement}
It is worth noting that the language model cannot be relied upon to generate outputs that are factually accurate, reliable, and knowledge-based for medical domain.
Infusing knowledge- and fact-based information through prompt engineering and calibration may alleviate the issue.


\bibliography{emnlp2023}

\begin{thebibliography}{64}
\expandafter\ifx\csname natexlab\endcsname\relax\def\natexlab#1{#1}\fi

\bibitem[{Aghajanyan et~al.(2021)Aghajanyan, Gupta, and
  Zettlemoyer}]{intrinsic-low-rank}
Armen Aghajanyan, Sonal Gupta, and Luke Zettlemoyer. 2021.
\newblock \href {https://doi.org/10.18653/v1/2021.acl-long.568} {Intrinsic
  dimensionality explains the effectiveness of language model fine-tuning}.
\newblock In \emph{Proceedings of the 59th Annual Meeting of the Association
  for Computational Linguistics and the 11th International Joint Conference on
  Natural Language Processing (Volume 1: Long Papers)}, pages 7319--7328,
  Online. Association for Computational Linguistics.

\bibitem[{Beltagy et~al.(2020)Beltagy, Peters, and Cohan}]{longformer}
Iz~Beltagy, Matthew~E. Peters, and Arman Cohan. 2020.
\newblock \href {https://arxiv.org/abs/2004.05150} {Longformer: The
  long-document transformer}.
\newblock \emph{arXiv preprint arXiv: 2004.05150}.

\bibitem[{Ben~Zaken et~al.(2022)Ben~Zaken, Goldberg, and Ravfogel}]{bitfit}
Elad Ben~Zaken, Yoav Goldberg, and Shauli Ravfogel. 2022.
\newblock \href {https://doi.org/10.18653/v1/2022.acl-short.1} {{B}it{F}it:
  Simple parameter-efficient fine-tuning for transformer-based masked
  language-models}.
\newblock In \emph{Proceedings of the 60th Annual Meeting of the Association
  for Computational Linguistics (Volume 2: Short Papers)}, pages 1--9, Dublin,
  Ireland. Association for Computational Linguistics.

\bibitem[{Brown et~al.(2020)Brown, Mann, Ryder, Subbiah, Kaplan, Dhariwal,
  Neelakantan, Shyam, Sastry, Askell, Agarwal, Herbert-Voss, Krueger, Henighan,
  Child, Ramesh, Ziegler, Wu, Winter, Hesse, Chen, Sigler, Litwin, Gray, Chess,
  Clark, Berner, McCandlish, Radford, Sutskever, and Amodei}]{gpt3-model}
Tom Brown, Benjamin Mann, Nick Ryder, Melanie Subbiah, Jared~D Kaplan, Prafulla
  Dhariwal, Arvind Neelakantan, Pranav Shyam, Girish Sastry, Amanda Askell,
  Sandhini Agarwal, Ariel Herbert-Voss, Gretchen Krueger, Tom Henighan, Rewon
  Child, Aditya Ramesh, Daniel Ziegler, Jeffrey Wu, Clemens Winter, Chris
  Hesse, Mark Chen, Eric Sigler, Mateusz Litwin, Scott Gray, Benjamin Chess,
  Jack Clark, Christopher Berner, Sam McCandlish, Alec Radford, Ilya Sutskever,
  and Dario Amodei. 2020.
\newblock \href
  {https://proceedings.neurips.cc/paper_files/paper/2020/file/1457c0d6bfcb4967418bfb8ac142f64a-Paper.pdf}
  {Language models are few-shot learners}.
\newblock In \emph{Advances in Neural Information Processing Systems
  (NeurIPS)}, volume~33, pages 1877--1901. Curran Associates, Inc.

\bibitem[{Child et~al.(2019)Child, Gray, Radford, and
  Sutskever}]{sparse-Transformer}
Rewon Child, Scott Gray, Alec Radford, and Ilya Sutskever. 2019.
\newblock \href {https://arxiv.org/abs/1904.10509} {Generating long sequences
  with sparse transformers}.
\newblock \emph{arXiv preprint arXiv: 1904.10509}.

\bibitem[{Chowdhery et~al.(2022)Chowdhery, Narang, Devlin, Bosma, Mishra,
  Roberts, Barham, Chung, Sutton, Gehrmann, Schuh, Shi, Tsvyashchenko, Maynez,
  Rao, Barnes, Tay, Shazeer, Prabhakaran, Reif, Du, Hutchinson, Pope, Bradbury,
  Austin, Isard, Gur-Ari, Yin, Duke, Levskaya, Ghemawat, Dev, Michalewski,
  Garcia, Misra, Robinson, Fedus, Zhou, Ippolito, Luan, Lim, Zoph, Spiridonov,
  Sepassi, Dohan, Agrawal, Omernick, Dai, Pillai, Pellat, Lewkowycz, Moreira,
  Child, Polozov, Lee, Zhou, Wang, Saeta, Diaz, Firat, Catasta, Wei,
  Meier-Hellstern, Eck, Dean, Petrov, and Fiedel}]{palm-model}
Aakanksha Chowdhery, Sharan Narang, Jacob Devlin, Maarten Bosma, Gaurav Mishra,
  Adam Roberts, Paul Barham, Hyung~Won Chung, Charles Sutton, Sebastian
  Gehrmann, Parker Schuh, Kensen Shi, Sasha Tsvyashchenko, Joshua Maynez,
  Abhishek Rao, Parker Barnes, Yi~Tay, Noam Shazeer, Vinodkumar Prabhakaran,
  Emily Reif, Nan Du, Ben Hutchinson, Reiner Pope, James Bradbury, Jacob
  Austin, Michael Isard, Guy Gur-Ari, Pengcheng Yin, Toju Duke, Anselm
  Levskaya, Sanjay Ghemawat, Sunipa Dev, Henryk Michalewski, Xavier Garcia,
  Vedant Misra, Kevin Robinson, Liam Fedus, Denny Zhou, Daphne Ippolito, David
  Luan, Hyeontaek Lim, Barret Zoph, Alexander Spiridonov, Ryan Sepassi, David
  Dohan, Shivani Agrawal, Mark Omernick, Andrew~M. Dai,
  Thanumalayan~Sankaranarayana Pillai, Marie Pellat, Aitor Lewkowycz, Erica
  Moreira, Rewon Child, Oleksandr Polozov, Katherine Lee, Zongwei Zhou, Xuezhi
  Wang, Brennan Saeta, Mark Diaz, Orhan Firat, Michele Catasta, Jason Wei,
  Kathy Meier-Hellstern, Douglas Eck, Jeff Dean, Slav Petrov, and Noah Fiedel.
  2022.
\newblock \href {https://arxiv.org/abs/2204.02311} {Palm: Scaling language
  modeling with pathways}.
\newblock \emph{arXiv preprint arXiv: 2204.02311}.

\bibitem[{Dettmers et~al.(2022)Dettmers, Lewis, Belkada, and
  Zettlemoyer}]{quantization-llm-8bit}
Tim Dettmers, Mike Lewis, Younes Belkada, and Luke Zettlemoyer. 2022.
\newblock \href
  {https://proceedings.neurips.cc/paper_files/paper/2022/file/c3ba4962c05c49636d4c6206a97e9c8a-Paper-Conference.pdf}
  {Gpt3.int8(): 8-bit matrix multiplication for transformers at scale}.
\newblock In \emph{Advances in Neural Information Processing Systems
  (NeurIPS)}, volume~35, pages 30318--30332. Curran Associates, Inc.

\bibitem[{Devlin et~al.(2019)Devlin, Chang, Lee, and Toutanova}]{bert-model}
Jacob Devlin, Ming-Wei Chang, Kenton Lee, and Kristina Toutanova. 2019.
\newblock {BERT}: Pre-training of deep bidirectional transformers for language
  understanding.
\newblock In \emph{Proceedings of the 2019 Conference of the North {A}merican
  Chapter of the Association for Computational Linguistics: Human Language
  Technologies (NAACL)}, pages 4171--4186, Minneapolis, Minnesota.

\bibitem[{Ding et~al.(2023)Ding, Qin, Yang, Wei, Yang, Su, Hu, Chen, Chan,
  Chen, Yi, Zhao, Wang, Liu, Zheng, Chen, Liu, Tang, Li, and
  Sun}]{delta-tuning}
Ning Ding, Yujia Qin, Guang Yang, Fuchao Wei, Zonghan Yang, Yusheng Su,
  Shengding Hu, Yulin Chen, Chi-Min Chan, Weize Chen, Jing Yi, Weilin Zhao,
  Xiaozhi Wang, Zhiyuan Liu, Hai-Tao Zheng, Jianfei Chen, Yang Liu, Jie Tang,
  Juanzi Li, and Maosong Sun. 2023.
\newblock \href {https://doi.org/https://doi.org/10.1038/s42256-023-00626-4}
  {Parameter-efficient fine-tuning of large-scale pre-trained language models}.
\newblock \emph{Nature Machine Intelligence}, 5:220--235.

\bibitem[{Fan et~al.(2020)Fan, Grave, and Joulin}]{layerdrop}
Angela Fan, Edouard Grave, and Armand Joulin. 2020.
\newblock \href {https://openreview.net/forum?id=SylO2yStDr} {Reducing
  transformer depth on demand with structured dropout}.
\newblock In \emph{International Conference on Learning Representations
  (ICLR)}.

\bibitem[{Goyal et~al.(2020)Goyal, Choudhury, Raje, Chakaravarthy, Sabharwal,
  and Verma}]{power-bert}
Saurabh Goyal, Anamitra~Roy Choudhury, Saurabh Raje, Venkatesan Chakaravarthy,
  Yogish Sabharwal, and Ashish Verma. 2020.
\newblock \href {https://proceedings.mlr.press/v119/goyal20a.html}
  {{P}o{WER}-{BERT}: Accelerating {BERT} inference via progressive word-vector
  elimination}.
\newblock In \emph{Proceedings of the International Conference on Machine
  Learning (ICML)}, volume 119 of \emph{Proceedings of Machine Learning
  Research}, pages 3690--3699. PMLR.

\bibitem[{Gu et~al.(2022)Gu, Han, Liu, and Huang}]{pretrain-prompt-tuning}
Yuxian Gu, Xu~Han, Zhiyuan Liu, and Minlie Huang. 2022.
\newblock \href {https://doi.org/10.18653/v1/2022.acl-long.576} {{PPT}:
  Pre-trained prompt tuning for few-shot learning}.
\newblock In \emph{Proceedings of the 60th Annual Meeting of the Association
  for Computational Linguistics (Volume 1: Long Papers)}, pages 8410--8423,
  Dublin, Ireland. Association for Computational Linguistics.

\bibitem[{Guo et~al.(2021)Guo, Rush, and Kim}]{diff-pruning}
Demi Guo, Alexander Rush, and Yoon Kim. 2021.
\newblock \href {https://doi.org/10.18653/v1/2021.acl-long.378}
  {Parameter-efficient transfer learning with diff pruning}.
\newblock In \emph{Proceedings of the 59th Annual Meeting of the Association
  for Computational Linguistics and the 11th International Joint Conference on
  Natural Language Processing (Volume 1: Long Papers)}, pages 4884--4896,
  Online. Association for Computational Linguistics.

\bibitem[{Guo et~al.(2022)Guo, Ainslie, Uthus, Ontanon, Ni, Sung, and
  Yang}]{longt5}
Mandy Guo, Joshua Ainslie, David Uthus, Santiago Ontanon, Jianmo Ni, Yun-Hsuan
  Sung, and Yinfei Yang. 2022.
\newblock \href {https://doi.org/10.18653/v1/2022.findings-naacl.55}
  {{L}ong{T}5: {E}fficient text-to-text transformer for long sequences}.
\newblock In \emph{Findings of the Association for Computational Linguistics:
  NAACL 2022}, pages 724--736, Seattle, United States. Association for
  Computational Linguistics.

\bibitem[{He et~al.(2022)He, Chen, Xie, Li, Doll\'ar, and
  Girshick}]{mae-linear-probing}
Kaiming He, Xinlei Chen, Saining Xie, Yanghao Li, Piotr Doll\'ar, and Ross
  Girshick. 2022.
\newblock Masked autoencoders are scalable vision learners.
\newblock In \emph{Proceedings of the IEEE/CVF Conference on Computer Vision
  and Pattern Recognition (CVPR)}, pages 16000--16009.

\bibitem[{Hedegaard et~al.(2022)Hedegaard, Alok, Jose, and Iosifidis}]{splora}
Lukas Hedegaard, Aman Alok, Juby Jose, and Alexandros Iosifidis. 2022.
\newblock \href {https://arxiv.org/abs/2211.10155} {Structured pruning
  adapters}.
\newblock \emph{arXiv preprint arXiv: 2211.10155}.

\bibitem[{Hinton et~al.(2019)Hinton, Vinyals, and Dean}]{distill-model}
Geoffrey Hinton, Oriol Vinyals, and Jeff Dean. 2019.
\newblock \href {https://arxiv.org/abs/1503.02531} {Distilling the knowledge in
  a neural network}.
\newblock \emph{arXiv preprint arXiv: 1503.02531}.

\bibitem[{Hou et~al.(2022)Hou, Pang, Zhou, Wu, Song, Song, and
  Zhou}]{token-dropping}
Le~Hou, Richard~Yuanzhe Pang, Tianyi Zhou, Yuexin Wu, Xinying Song, Xiaodan
  Song, and Denny Zhou. 2022.
\newblock \href {https://doi.org/10.18653/v1/2022.acl-long.262} {Token dropping
  for efficient {BERT} pretraining}.
\newblock In \emph{Proceedings of the 60th Annual Meeting of the Association
  for Computational Linguistics (Volume 1: Long Papers)}, pages 3774--3784,
  Dublin, Ireland. Association for Computational Linguistics.

\bibitem[{Houlsby et~al.(2019)Houlsby, Giurgiu, Jastrzebski, Morrone,
  de~Laroussilhe, Gesmundo, Attariyan, and Gelly}]{adapter-model}
Neil Houlsby, Andrei Giurgiu, Stanislaw Jastrzebski, Bruna Morrone, Quentin
  de~Laroussilhe, Andrea Gesmundo, Mona Attariyan, and Sylvain Gelly. 2019.
\newblock Parameter-efficient transfer learning for nlp.
\newblock In \emph{Proceedings of the International Conference on Machine
  Learning (ICML)}.

\bibitem[{Hu et~al.(2022)Hu, yelong shen, Wallis, Allen-Zhu, Li, Wang, Wang,
  and Chen}]{lora-model}
Edward~J Hu, yelong shen, Phillip Wallis, Zeyuan Allen-Zhu, Yuanzhi Li, Shean
  Wang, Lu~Wang, and Weizhu Chen. 2022.
\newblock \href {https://openreview.net/forum?id=nZeVKeeFYf9} {Lo{RA}: Low-rank
  adaptation of large language models}.
\newblock In \emph{International Conference on Learning Representations
  (ICLR)}.

\bibitem[{Jacob et~al.(2018)Jacob, Kligys, Chen, Zhu, Tang, Howard, Adam, and
  Kalenichenko}]{quantization-2018}
Benoit Jacob, Skirmantas Kligys, Bo~Chen, Menglong Zhu, Matthew Tang, Andrew
  Howard, Hartwig Adam, and Dmitry Kalenichenko. 2018.
\newblock \href {https://doi.org/10.1109/CVPR.2018.00286} {Quantization and
  training of neural networks for efficient integer-arithmetic-only inference}.
\newblock In \emph{Proceedings of the IEEE Conference on Computer Vision and
  Pattern Recognition (CVPR)}.

\bibitem[{Jiao et~al.(2020)Jiao, Yin, Shang, Jiang, Chen, Li, Wang, and
  Liu}]{tinybert}
Xiaoqi Jiao, Yichun Yin, Lifeng Shang, Xin Jiang, Xiao Chen, Linlin Li, Fang
  Wang, and Qun Liu. 2020.
\newblock \href {https://doi.org/10.18653/v1/2020.findings-emnlp.372}
  {{T}iny{BERT}: Distilling {BERT} for natural language understanding}.
\newblock In \emph{Findings of the Association for Computational Linguistics:
  EMNLP 2020}, pages 4163--4174, Online. Association for Computational
  Linguistics.

\bibitem[{Johnson et~al.(2023{\natexlab{a}})Johnson, Bulgarelli, Pollard,
  Horng, Celi, and Mark}]{mimic-iv-cite}
Alistair Johnson, Lucas Bulgarelli, Tom Pollard, Steven Horng, Leo~Anthony
  Celi, and Roger Mark. 2023{\natexlab{a}}.
\newblock \href {https://doi.org/https://doi.org/10.1038/s41597-022-01899-x}
  {Mimic-iv}.
\newblock \emph{PhysioNet}.

\bibitem[{Johnson et~al.(2023{\natexlab{b}})Johnson, Bulgarelli, Shen, Gayles,
  Shammout, Horng, Pollard, Hao, Moody, Gow, wei H.~Lehman, Celi, and
  Mark}]{mimic-iv-paper}
Alistair Johnson, Lucas Bulgarelli, Lu~Shen, Alvin Gayles, Ayad Shammout,
  Steven Horng, Tom~J. Pollard, Sicheng Hao, Benjamin Moody, Brian Gow, Li~wei
  H.~Lehman, Leo~A. Celi, and Roger~G. Mark. 2023{\natexlab{b}}.
\newblock \href {https://doi.org/https://doi.org/10.1038/s41597-023-02136-9}
  {Mimic-iv, a freely accessible electronic health record dataset}.
\newblock \emph{Scientific Data}, 10.

\bibitem[{Johnson et~al.(2023{\natexlab{c}})Johnson, Pollard, Horng, Celi, and
  Mark}]{mimic-iv-note-cite}
Alistair Johnson, Tom Pollard, Steven Horng, Leo~Anthony Celi, and Roger Mark.
  2023{\natexlab{c}}.
\newblock \href {https://doi.org/https://doi.org/10.13026/1n74-ne17}
  {Mimic-iv-note: Deidentified free-text clinical notes}.
\newblock \emph{PhysioNet}.

\bibitem[{Karimi~Mahabadi et~al.(2021)Karimi~Mahabadi, Henderson, and
  Ruder}]{compacter-model}
Rabeeh Karimi~Mahabadi, James Henderson, and Sebastian Ruder. 2021.
\newblock \href
  {https://proceedings.neurips.cc/paper_files/paper/2021/file/081be9fdff07f3bc808f935906ef70c0-Paper.pdf}
  {Compacter: Efficient low-rank hypercomplex adapter layers}.
\newblock In \emph{Advances in Neural Information Processing Systems
  (NeurIPS)}, pages 1022--1035.

\bibitem[{Kim et~al.(2022)Kim, Shen, Thorsley, Gholami, Kwon, Hassoun, and
  Keutzer}]{learned-token-pruning}
Sehoon Kim, Sheng Shen, David Thorsley, Amir Gholami, Woosuk Kwon, Joseph
  Hassoun, and Kurt Keutzer. 2022.
\newblock \href {https://doi.org/10.1145/3534678.3539260} {Learned token
  pruning for transformers}.
\newblock In \emph{Proceedings of the 28th ACM SIGKDD Conference on Knowledge
  Discovery and Data Mining}, KDD '22, page 784–794, New York, NY, USA.
  Association for Computing Machinery.

\bibitem[{Kim and Rush(2016)}]{seq-knowledge-distill}
Yoon Kim and Alexander~M. Rush. 2016.
\newblock \href {https://doi.org/10.18653/v1/D16-1139} {Sequence-level
  knowledge distillation}.
\newblock In \emph{Proceedings of the 2016 Conference on Empirical Methods in
  Natural Language Processing (EMNLP)}, pages 1317--1327, Austin, Texas.
  Association for Computational Linguistics.

\bibitem[{Kitaev et~al.(2020)Kitaev, Kaiser, and Levskaya}]{reformer}
Nikita Kitaev, Lukasz Kaiser, and Anselm Levskaya. 2020.
\newblock \href {https://openreview.net/forum?id=rkgNKkHtvB} {Reformer: The
  efficient transformer}.
\newblock In \emph{International Conference on Learning Representations
  (ICLR)}.

\bibitem[{Laban et~al.(2022)Laban, Schnabel, Bennett, and
  Hearst}]{summac-metric}
Philippe Laban, Tobias Schnabel, Paul~N. Bennett, and Marti~A. Hearst. 2022.
\newblock \href {https://doi.org/https://doi.org/10.1162/tacl_a_00453} {Summac:
  Re-visiting nli-based models for inconsistency detection in summarization}.
\newblock \emph{Transactions of the Association for Computational Linguistics},
  10:163--177.

\bibitem[{Lagunas et~al.(2021)Lagunas, Charlaix, Sanh, and
  Rush}]{block-pruning}
Fran{\c{c}}ois Lagunas, Ella Charlaix, Victor Sanh, and Alexander Rush. 2021.
\newblock \href {https://doi.org/10.18653/v1/2021.emnlp-main.829} {Block
  pruning for faster transformers}.
\newblock In \emph{Proceedings of the 2021 Conference on Empirical Methods in
  Natural Language Processing (EMNLP)}, pages 10619--10629, Online and Punta
  Cana, Dominican Republic. Association for Computational Linguistics.

\bibitem[{Lee et~al.(2019)Lee, Tang, and Lin}]{elsa-freezing}
Jaejun Lee, Raphael Tang, and Jimmy Lin. 2019.
\newblock \href {https://arxiv.org/abs/1911.03090} {What would elsa do?
  freezing layers during transformer fine-tuning}.
\newblock \emph{arXiv preprint arXiv: 1911.03090}.

\bibitem[{Lei et~al.(2023)Lei, Bai, Brahma, Ainslie, Lee, Zhou, Du, Zhao, Wu,
  Li, Zhang, and Chang}]{conditional-adpater}
Tao Lei, Junwen Bai, Siddhartha Brahma, Joshua Ainslie, Kenton Lee, Yanqi Zhou,
  Nan Du, Vincent~Y. Zhao, Yuexin Wu, Bo~Li, Yu~Zhang, and Ming-Wei Chang.
  2023.
\newblock Conditional adapters: Parameter-efficient transfer learning with fast
  inference.
\newblock \emph{arXiv preprint arXiv: 2304.04947}.

\bibitem[{Lester et~al.(2021)Lester, Al-Rfou, and
  Constant}]{prompt-tuning-largescale}
Brian Lester, Rami Al-Rfou, and Noah Constant. 2021.
\newblock \href {https://doi.org/10.18653/v1/2021.emnlp-main.243} {The power of
  scale for parameter-efficient prompt tuning}.
\newblock In \emph{Proceedings of the 2021 Conference on Empirical Methods in
  Natural Language Processing (EMNLP)}, pages 3045--3059, Online and Punta
  Cana, Dominican Republic. Association for Computational Linguistics.

\bibitem[{Lewis et~al.(2020)Lewis, Liu, Goyal, Ghazvininejad, Mohamed, Levy,
  Stoyanov, and Zettlemoyer}]{bart-model}
Mike Lewis, Yinhan Liu, Naman Goyal, Marjan Ghazvininejad, Abdelrahman Mohamed,
  Omer Levy, Veselin Stoyanov, and Luke Zettlemoyer. 2020.
\newblock \href {https://doi.org/https://doi.org/10.18653/v1/2020.acl-main.703}
  {{BART}: Denoising sequence-to-sequence pre-training for natural language
  generation, translation, and comprehension.}
\newblock In \emph{Proceedings of the 58th Annual Meeting of the Association
  for Computational Linguistics (ACL)}, pages 7871--7880, Online.

\bibitem[{Li et~al.(2018)Li, Farkhoor, Liu, and Yosinski}]{intrinsic-dim}
Chunyuan Li, Heerad Farkhoor, Rosanne Liu, and Jason Yosinski. 2018.
\newblock \href {https://openreview.net/forum?id=ryup8-WCW} {Measuring the
  intrinsic dimension of objective landscapes}.
\newblock In \emph{International Conference on Learning Representations
  (ICLR)}.

\bibitem[{Li and Liang(2021)}]{prefix-tuning}
Xiang~Lisa Li and Percy Liang. 2021.
\newblock \href {https://doi.org/10.18653/v1/2021.acl-long.353} {Prefix-tuning:
  Optimizing continuous prompts for generation}.
\newblock In \emph{Proceedings of the 59th Annual Meeting of the Association
  for Computational Linguistics and the 11th International Joint Conference on
  Natural Language Processing (Volume 1: Long Papers)}, pages 4582--4597,
  Online. Association for Computational Linguistics.

\bibitem[{Li et~al.(2023)Li, Li, Zhang, Dan, and Zhang}]{chatdoctor-paper}
Yunxiang Li, Zihan Li, Kai Zhang, Ruilong Dan, and You Zhang. 2023.
\newblock \href {http://arxiv.org/abs/2303.14070} {Chatdoctor: A medical chat
  model fine-tuned on llama model using medical domain knowledge}.

\bibitem[{Lin(2004)}]{rouge-metric}
Chin-Yew Lin. 2004.
\newblock \href {https://aclanthology.org/W04-1013} {{ROUGE}: A package for
  automatic evaluation of summaries.}
\newblock In \emph{Proceedings of the 42nd Annual Meeting of the Association
  for Computational Linguistics (ACL)}, pages 74--81.

\bibitem[{Liu et~al.(2022)Liu, Ji, Fu, Tam, Du, Yang, and Tang}]{p-tuning-v2}
Xiao Liu, Kaixuan Ji, Yicheng Fu, Weng Tam, Zhengxiao Du, Zhilin Yang, and Jie
  Tang. 2022.
\newblock \href {https://doi.org/10.18653/v1/2022.acl-short.8} {{P}-tuning:
  Prompt tuning can be comparable to fine-tuning across scales and tasks}.
\newblock In \emph{Proceedings of the 60th Annual Meeting of the Association
  for Computational Linguistics (Volume 2: Short Papers)}, pages 61--68,
  Dublin, Ireland. Association for Computational Linguistics.

\bibitem[{Liu et~al.(2019)Liu, Zheng, Du, Ding, Qian, Yang, and
  Tang}]{p-tuning}
Xiao Liu, Yanan Zheng, Zhengxiao Du, Ming Ding, Yujie Qian, Zhilin Yang, and
  Jie Tang. 2019.
\newblock \href {https://arxiv.org/abs/2103.10385} {Gpt understands, too}.
\newblock \emph{arXiv preprint arXiv: 2103.10385}.

\bibitem[{Loshchilov and Hutter(2019)}]{adamW}
Ilya Loshchilov and Frank Hutter. 2019.
\newblock \href {https://openreview.net/forum?id=Bkg6RiCqY7} {Decoupled weight
  decay regularization}.
\newblock In \emph{International Conference on Learning Representations
  (ICLR)}.

\bibitem[{Papineni et~al.(2002)Papineni, Roukos, Ward, and Zhu}]{bleu-metric}
Kishore Papineni, Salim Roukos, Todd Ward, and Wei-Jing Zhu. 2002.
\newblock \href {https://doi.org/https://doi.org/10.3115/1073083.1073135}
  {{B}leu: a method for automatic evaluation of machine translation.}
\newblock In \emph{Proceedings of the 40th Annual Meeting of the Association
  for Computational Linguistics (ACL)}, pages 311--318.

\bibitem[{Pfeiffer et~al.(2021)Pfeiffer, Kamath, R{\"u}ckl{\'e}, Cho, and
  Gurevych}]{adapterfusion-model}
Jonas Pfeiffer, Aishwarya Kamath, Andreas R{\"u}ckl{\'e}, Kyunghyun Cho, and
  Iryna Gurevych. 2021.
\newblock \href {https://doi.org/https://doi.org/10.18653/v1/2021.eacl-main.39}
  {{A}dapter{F}usion: Non-destructive task composition for transfer learning}.
\newblock In \emph{Proceedings of the 16th Conference of the European Chapter
  of the Association for Computational Linguistics: Main Volume}, pages
  487--503, Online. Association for Computational Linguistics.

\bibitem[{Pfeiffer et~al.(2020)Pfeiffer, Vuli{\'c}, Gurevych, and
  Ruder}]{mad-x-model}
Jonas Pfeiffer, Ivan Vuli{\'c}, Iryna Gurevych, and Sebastian Ruder. 2020.
\newblock \href
  {https://doi.org/https://doi.org/10.18653/v1/2020.emnlp-main.617} {{MAD-X}:
  {A}n {A}dapter-{B}ased {F}ramework for {M}ulti-{T}ask {C}ross-{L}ingual
  {T}ransfer}.
\newblock In \emph{Proceedings of the 2020 Conference on Empirical Methods in
  Natural Language Processing (EMNLP)}, pages 7654--7673, Online. Association
  for Computational Linguistics.

\bibitem[{Radford et~al.(2018)Radford, Narasimhan, Salimans, and
  Sutskever}]{gpt-1}
Alec Radford, Karthik Narasimhan, Tim Salimans, and Ilya Sutskever. 2018.
\newblock \href
  {https://s3-us-west-2.amazonaws.com/openai-assets/research-covers/language-unsupervised/language_understanding_paper.pdf}
  {Improving language understanding by generative pre-training}.

\bibitem[{Raffel et~al.(2020)Raffel, Shazeer, Roberts, Lee, Narang, Matena,
  Zhou, Li, and Liu}]{t5-model}
Colin Raffel, Noam Shazeer, Adam Roberts, Katherine Lee, Sharan Narang, Michael
  Matena, Yanqi Zhou, Wei Li, and Peter~J. Liu. 2020.
\newblock \href {http://jmlr.org/papers/v21/20-074.html} {Exploring the limits
  of transfer learning with a unified text-to-text transformer}.
\newblock \emph{Journal of Machine Learning Research}, 21(140):1--67.

\bibitem[{Rebuffi et~al.(2017)Rebuffi, Bilen, and Vedaldi}]{residual-adapters}
Sylvestre-Alvise Rebuffi, Hakan Bilen, and Andrea Vedaldi. 2017.
\newblock \href
  {https://proceedings.neurips.cc/paper_files/paper/2017/file/e7b24b112a44fdd9ee93bdf998c6ca0e-Paper.pdf}
  {Learning multiple visual domains with residual adapters}.
\newblock In \emph{Advances in Neural Information Processing Systems
  (NeurIPS)}.

\bibitem[{R{\"u}ckl{\'e} et~al.(2021)R{\"u}ckl{\'e}, Geigle, Glockner, Beck,
  Pfeiffer, Reimers, and Gurevych}]{adapterdrop-model}
Andreas R{\"u}ckl{\'e}, Gregor Geigle, Max Glockner, Tilman Beck, Jonas
  Pfeiffer, Nils Reimers, and Iryna Gurevych. 2021.
\newblock \href
  {https://doi.org/https://doi.org/10.18653/v1/2021.emnlp-main.626}
  {{AdapterDrop}: {O}n the efficiency of adapters in transformers}.
\newblock In \emph{Proceedings of the 2021 Conference on Empirical Methods in
  Natural Language Processing (EMNLP)}, pages 7930--7946, Online and Punta
  Cana, Dominican Republic. Association for Computational Linguistics.

\bibitem[{Sajjad et~al.(2023)Sajjad, Dalvi, Durrani, and
  Nakov}]{layer-dropping}
Hassan Sajjad, Fahim Dalvi, Nadir Durrani, and Preslav Nakov. 2023.
\newblock \href {https://doi.org/https://doi.org/10.1016/j.csl.2022.101429} {On
  the effect of dropping layers of pre-trained transformer models}.
\newblock \emph{Computer Speech \& Language}, 77:101429.

\bibitem[{Sanh et~al.(2019)Sanh, Debut, Chaumond, and Wolf}]{distill-bert}
Victor Sanh, Lysandre Debut, Julien Chaumond, and Thomas Wolf. 2019.
\newblock \href {https://arxiv.org/abs/1910.01108} {Distilbert, a distilled
  version of bert: smaller, faster, cheaper and lighter}.
\newblock \emph{arXiv preprint arXiv: 1910.01108}.

\bibitem[{Sanh et~al.(2020)Sanh, Wolf, and Rush}]{movement-pruning}
Victor Sanh, Thomas Wolf, and Alexander Rush. 2020.
\newblock \href
  {https://proceedings.neurips.cc/paper_files/paper/2020/file/eae15aabaa768ae4a5993a8a4f4fa6e4-Paper.pdf}
  {Movement pruning: Adaptive sparsity by fine-tuning}.
\newblock In \emph{Advances in Neural Information Processing Systems
  (NeurIPS)}, volume~33, pages 20378--20389. Curran Associates, Inc.

\bibitem[{Vaswani et~al.(2017)Vaswani, Shazeer, Parmar, Uszkoreit, Jones,
  Gomez, Kaiser, and Polosukhin}]{Transformer-model}
Ashish Vaswani, Noam Shazeer, Niki Parmar, Jakob Uszkoreit, Llion Jones,
  Aidan~N. Gomez, Lukasz Kaiser, and Illia Polosukhin. 2017.
\newblock Attention is all you need.
\newblock In \emph{Advances in Neural Information Processing Systems
  (NeurIPS)}.

\bibitem[{Wang et~al.(2023)Wang, Yang, and Sun}]{prune-lora}
Guorun Wang, Jun Yang, and Yaoru Sun. 2023.
\newblock \href {https://arxiv.org/abs/2303.14704} {Task-oriented
  memory-efficient pruning-adapter}.
\newblock \emph{arXiv preprint arXiv: 2303.14704}.

\bibitem[{Wang et~al.(2018)Wang, Choi, Brand, Chen, and
  Gopalakrishnan}]{quantization-training}
Naigang Wang, Jungwook Choi, Daniel Brand, Chia-Yu Chen, and Kailash
  Gopalakrishnan. 2018.
\newblock \href
  {https://proceedings.neurips.cc/paper_files/paper/2018/file/335d3d1cd7ef05ec77714a215134914c-Paper.pdf}
  {Training deep neural networks with 8-bit floating point numbers}.
\newblock In \emph{Advances in Neural Information Processing Systems
  (NeurIPS)}, volume~31. Curran Associates, Inc.

\bibitem[{Wang et~al.(2021)Wang, Tang, Duan, Wei, Huang, Ji, Cao, Jiang, and
  Zhou}]{k-adapter-model}
Ruize Wang, Duyu Tang, Nan Duan, Zhongyu Wei, Xuanjing Huang, Jianshu Ji,
  Guihong Cao, Daxin Jiang, and Ming Zhou. 2021.
\newblock \href
  {https://doi.org/https://doi.org/10.18653/v1/2021.findings-acl.121}
  {{K-Adapter}: {I}nfusing {K}nowledge into {P}re-{T}rained {M}odels with
  {A}dapters}.
\newblock In \emph{Findings of the Association for Computational Linguistics:
  ACL-IJCNLP 2021}, pages 1405--1418, Online. Association for Computational
  Linguistics.

\bibitem[{Wang et~al.(2022)Wang, Agarwal, Mukherjee, Liu, Gao, Awadallah, and
  Gao}]{adamix}
Yaqing Wang, Sahaj Agarwal, Subhabrata Mukherjee, Xiaodong Liu, Jing Gao,
  Ahmed~Hassan Awadallah, and Jianfeng Gao. 2022.
\newblock \href {https://aclanthology.org/2022.emnlp-main.388} {{A}da{M}ix:
  Mixture-of-adaptations for parameter-efficient model tuning}.
\newblock In \emph{Proceedings of the 2022 Conference on Empirical Methods in
  Natural Language Processing (EMNLP)}, pages 5744--5760, Abu Dhabi, United
  Arab Emirates. Association for Computational Linguistics.

\bibitem[{Xia et~al.(2022)Xia, Zhong, and Chen}]{cofi-model}
Mengzhou Xia, Zexuan Zhong, and Danqi Chen. 2022.
\newblock \href {https://doi.org/10.18653/v1/2022.acl-long.107} {Structured
  pruning learns compact and accurate models}.
\newblock In \emph{Proceedings of the 60th Annual Meeting of the Association
  for Computational Linguistics (Volume 1: Long Papers)}, pages 1513--1528,
  Dublin, Ireland. Association for Computational Linguistics.

\bibitem[{Yu et~al.(2023)Yu, Chang, Liu, Tian, and Chen}]{uni-peft-model}
Bruce Yu, Jianlong Chang, Lingbo Liu, Qi~Tian, and Chang~Wen Chen. 2023.
\newblock \href {https://openreview.net/forum?id=ti6fH3EhFkv} {Towards a
  unified view on visual parameter-efficient transfer learning}.
\newblock In \emph{International Conference on Learning Representations
  (ICLR)}.

\bibitem[{Zeng et~al.(2020)Zeng, Yang, Ju, Yang, Wang, Zhang, Zhou, Zeng, Dong,
  Zhang, Fang, Zhu, Chen, and Xie}]{meddialog-paper}
Guangtao Zeng, Wenmian Yang, Zeqian Ju, Yue Yang, Sicheng Wang, Ruisi Zhang,
  Meng Zhou, Jiaqi Zeng, Xiangyu Dong, Ruoyu Zhang, Hongchao Fang, Penghui Zhu,
  Shu Chen, and Pengtao Xie. 2020.
\newblock \href
  {https://doi.org/https://doi.org/10.18653/v1/2020.emnlp-main.743}
  {{M}ed{D}ialog: Large-scale medical dialogue datasets}.
\newblock In \emph{Proceedings of the 2020 Conference on Empirical Methods in
  Natural Language Processing (EMNLP)}, pages 9241--9250.

\bibitem[{Zhang and He(2020)}]{progressive-pruning}
Minjia Zhang and Yuxiong He. 2020.
\newblock \href
  {https://proceedings.neurips.cc/paper_files/paper/2020/file/a1140a3d0df1c81e24ae954d935e8926-Paper.pdf}
  {Accelerating training of transformer-based language models with progressive
  layer dropping}.
\newblock In \emph{Advances in Neural Information Processing Systems},
  volume~33, pages 14011--14023. Curran Associates, Inc.

\bibitem[{Zhang et~al.(2023)Zhang, Chen, Bukharin, He, Cheng, Chen, and
  Zhao}]{adalora-model}
Qingru Zhang, Minshuo Chen, Alexander Bukharin, Pengcheng He, Yu~Cheng, Weizhu
  Chen, and Tuo Zhao. 2023.
\newblock \href {https://openreview.net/forum?id=lq62uWRJjiY} {Adaptive budget
  allocation for parameter-efficient fine-tuning}.
\newblock In \emph{International Conference on Learning Representations
  (ICLR)}.

\bibitem[{Zhang et~al.(2020)Zhang, Kishore, Wu, Weinberger, and
  Artzi}]{bertscore-metric}
Tianyi Zhang, Varsha Kishore, Felix Wu, Kilian~Q. Weinberger, and Yoav Artzi.
  2020.
\newblock \href {https://openreview.net/forum?id=SkeHuCVFDr} {Bertscore:
  Evaluating text generation with {BERT}}.
\newblock In \emph{International Conference on Learning Representations
  (ICLR)}.

\bibitem[{Zhao et~al.(2020)Zhao, Lin, Mi, Jaggi, and
  Sch{\"u}tze}]{masking-efficient}
Mengjie Zhao, Tao Lin, Fei Mi, Martin Jaggi, and Hinrich Sch{\"u}tze. 2020.
\newblock \href {https://doi.org/10.18653/v1/2020.emnlp-main.174} {Masking as
  an efficient alternative to finetuning for pretrained language models}.
\newblock In \emph{Proceedings of the 2020 Conference on Empirical Methods in
  Natural Language Processing (EMNLP)}, pages 2226--2241, Online. Association
  for Computational Linguistics.

\end{thebibliography}
\bibliographystyle{acl_natbib}

\newpage
\appendix

\definecolor{c_blue1}{RGB}{108, 155, 207}

\section{Appendix}

\subsection{Hyperparameter Analysis} \label{sec:appendix-hyper}
We show the experimental results of different hyperparameter settings of  $r$ for LoRA in Table~\ref{tab:hyper-alpha}.
\begin{table}[htbp]
  \centering
  \scalebox{0.6}{
    \begin{tblr}{colspec={lcrrcc}, rowsep=0.5pt, stretch=1.1, rows={ht=\baselineskip}, row{5, 9} = {bg=c_grey1}}
      \toprule
      Model                      & Speed  & Trained Params & Mem Used & R-1   & R-2   \\
      \hline
      BART-large                 & 100 \% & 406.2 M        & 34.56 GB & 43.46 & 25.05 \\
      $+$ LoRA$_{QV}$ ($r = 4$)  & 115 \% & 0.6 M          & 28.11 GB & 38.07 & 21.06 \\
      $+$ LoRA$_{QV}$ ($r = 8$)  & 115 \% & 1.2 M          & 28.12 GB & 39.64 & 22.23 \\
      $+$ LoRA$_{QV}$ ($r = 16$) & 115 \% & 2.4 M          & 28.15 GB & 40.32 & 22.66 \\
      $+$ LoRA$_{QV}$ ($r = 32$) & 114 \% & 4.7 M          & 28.20 GB & 40.88 & 23.10 \\

      $+$ LoRA$_{FF}$ ($r = 4$)  & 127 \% & 1.1 M          & 27.19 GB & 39.79 & 22.50 \\
      $+$ LoRA$_{FF}$ ($r = 8$)  & 127 \% & 2.0 M          & 27.21 GB & 40.73 & 23.19 \\
      $+$ LoRA$_{FF}$ ($r = 16$) & 125 \% & 4.0 M          & 27.25 GB & 41.60 & 23.77 \\
      $+$ LoRA$_{FF}$ ($r = 32$) & 119 \% & 7.9 M          & 27.32 GB & 42.24 & 24.19 \\
      \bottomrule
    \end{tblr}
  }
  \caption{Comparison of different rank $r$ for LoRA in BART-large on MIMIC-IV-discharge.}
  \label{tab:hyper-alpha}
\end{table}

\subsection{Machine-Generated Examples} \label{sec:appendix-example}
The following Tables \ref{tab:case-dis}, \ref{tab:case-rad}, \ref{tab:case-hcm} and \ref{tab:case-icliniq} show a few machine-generated examples of MIMIC-IV-discharge, MIMIC-IV-radiology, HealthCareMagic and iCliniq, respectively.
The index of a example in the test set is indicated by the number after the symbol \#.
Quantitative evaluations of the outputs are shown at the end of each row in blue.
\begin{table*}[htbp]
  \centering \renewcommand\arraystretch{1.0} \small
  \begin{tabular}{p{0.95\textwidth}}
    \hline
    \textbf{Source:}
    service: medicine.
    chief complaint:
    s/p cardiac arrest
      [...]
    history of present illness:
    \_\_\_ y/o f with h/o severe aortic stenosis, esrd on hd, paroxysmal
    atrial fibrillation, and ppm for lbbb/syncope presents following
    acute loc while at hd with pulselessness. patient underwent
    \_\_\_ minutes of cpr (no epi or defibrillation per ed report)
    before rosc was achieved.
      [...]
    patient was awake and talking and denied cp or sob, though she endorsed significant ongoing
    back pain, which has been a chronic issue. bp subsequently
    improved with initiation of levophed. cardiology was consulted
    due to concern for posterior stemi given ekg findings;
    [...]
    in the ed initial vitals were:
    [...]
    moderate to marked enlargement of the cardiac silhouette.
    multiple bilateral rib fractures.
      [...]
    past medical history:
    1. cardiac risk factors
    - esrd on the basis of hypertension, on hemodialysis with r av fistula
      [...]
    -nonobstructive coronary artery disease.
    -hypothyroidism.
    [...]
    echo:
    the left atrium is moderately dilated. there is moderate
    symmetric left ventricular hypertrophy. overall left ventricular
    systolic function is severely depressed (lvef = 25\%) secondary
    to akinesis of the inferior wall, hypokinesis of the posterior
    wall, and pacing-induced dyssynchrony. tissue doppler imaging
    suggests an increased left ventricular filling pressure
    (pcwp>18mmhg). doppler parameters are most consistent with grade
    iii/iv (severe) left ventricular diastolic dysfunction. the
    right ventricular free wall thickness is normal. right
    ventricular chamber size is normal with borderline normal free
    wall function. the aortic valve is not well seen.
      [...]
    there is moderate pulmonary artery systolic hypertension. there is no pericardial effusion.
      [...]
    hospital course:
    \_\_\_ y/o f with h/o severe aortic stenosis, esrd on hd, paroxysmal
    atrial fibrillation, and ppm for lbbb/syncope who presents
    following cardiac arrest, transferred to \_\_\_ for further
    management of hypotension associated with hd sessions and for
    post cardiac arrest care.
      [...]
    chronic issues:
    \#copd on home o2 (2l): not on any inhalers per patient. continue
    2l o2. continued on nebulizer treatment.
    \#chronic back pain: continued gabapentin and methadone
      [...]
    \\ \hline
    \textbf{Reference:}
    primary diagnosis:
    aortic stenosis.
    congestive heart failure.
    s/p cardiac arrest.
    hypotension.
    atrial fibrillation with rapid ventricular response.
    transiminitis.
    end stage renal disease on hemodialysis.
    rib fractures.
    seconadry diagnosis:
    chronic obstructive pulmonary disease.
    gastroesophageal reflux disease.
    chronic back pain.
    hypothyroidism.
    \\ \hline
    \textbf{BART-large:}
    primary diagnosis:
    cardiac arrest.
    pulmonary edema.
    severe aortic stenosis.
    end stage renal disease on hemodialysis.
    \\ \textcolor{c_blue1}{(R-1: 47.27; R-2: 30.18; BertScore: 85.08; SummaC: 41.49)}
    \\ \hline
    \textbf{BART-large $+$ L-Prun:}
    primary diagnosis:
    cardiac arrest.
    severe aortic stenosis.
    secondary diagnoses:
    atrial fibrillation.
    end stage renal disease on hemodialysis.
    \\ \textcolor{c_blue1}{(R-1: 49.12; R-2: 32.72; BertScore: 85.72; SummaC: 46.19)}
    \\ \hline
    \textbf{BART-large $+$ LoRA$_{FF}$:}
    primary diagnosis:
    cardiac arrest.
    severe aortic stenosis.
    secondary diagnoses:
    end stage renal disease.
    paroxysmal atrial fibrillation.
    \\ \textcolor{c_blue1}{(R-1: 42.85; R-2: 25.92; BertScore: 83.67; SummaC: 43.17)}
    \\ \hline
    \textbf{BART-large $+$ LoRA$_{FF}$ $+$ L-Prun:}
    primary diagnosis:
    non st elevation myocardial infarction.
    acute on chronic systolic heart failure.
    atrial fibrillation with rapid ventricular response.
    secondary diagnosis:
    end stage renal disease on hemodialysis.
    hypertension.
    hyperlipidemia.
    \\ \textcolor{c_blue1}{(R-1: 52.17; R-2: 35.82; BertScore: 85.92; SummaC: 36.88)}
    \\ \hline
    \textbf{T5-large:}
    primary diagnosis:
    cardiac arrest severe aortic stenosis.
    secondary diagnosis:
    esrd on hd.
    \\ \textcolor{c_blue1}{(R-1: 30.76; R-2: 12.00; BertScore: 79.84; SummaC: 55.46)}
    \\ \hline
    \textbf{T5-large $+$ L-Prun:}
    primary diagnosis:
    s/p cardiac arrest severe aortic stenosis.
    secondary diagnosis:
    esrd on hd paroxysmal atrial fibrillation.
    \\ \textcolor{c_blue1}{(R-1: 42.10; R-2: 21.81; BertScore: 81.08; SummaC: 42.15)}
    \\ \hline
    \textbf{T5-large $+$ LoRA$_{FF}$:}
    s/p cardiac arrest severe aortic stenosis end stage renal disease on hemodialysis paroxysmal atrial fibrillation.
    \\ \textcolor{c_blue1}{(R-1: 50.00; R-2: 37.03; BertScore: 83.95; SummaC: 45.61)}
    \\ \hline
    \textbf{T5-large $+$ LoRA$_{FF}$ $+$ L-Prun:}
    primary diagnosis:
    severe aortic stenosis.
    secondary diagnosis:
    end stage renal disease on hemodialysis paroxysmal atrial fibrillation.
    \\ \textcolor{c_blue1}{(R-1: 46.42; R-2: 29.62; BertScore: 83.03; SummaC: 45.67)}
    \\ \hline \\
  \end{tabular}
  \caption{Generated examples of MIMIC-IV-discharge. \# 32.}
  \label{tab:case-dis}
\end{table*}

\begin{table*}[htbp]
  \centering \renewcommand\arraystretch{1.0} \small
  \begin{tabular}{p{0.95\textwidth}}
    \hline
    \textbf{Source:}
    findings:
    neck, thoracic inlet, axillae, chest wall: the thyroid is notable for a large
    partially peripherally calcified nodule in the right lobe measuring up to 2.5
    x 1.9 cm (03:23) there is no axillary lymphadenopathy.  tiny supraclavicular
    lymph nodes are unchanged since the prior examinations (for example 03:20).
    the subcutaneous tissues of the chest wall are unremarkable.
    upper abdomen: please see report from dedicated ct of the abdomen and pelvis
    for subdiaphragmatic findings.
    mediastinum: there is no mediastinal lymphadenopathy.  the lower esophagus is
    mildly dilated.
    hila: there is no hilar lymphadenopathy.
    heart and pericardium: the heart is normal in size.  there is no pericardial
    effusion.
    pleura: no pleural effusion or pneumothorax
    lung:
    1. parenchyma:  re-demonstrated are several millimetric pulmonary nodules,
    unchanged since the most recent study, and decreased in size since the
    examination from \_\_\_.  a right lower lobe nodule (3:180) measures up to
    4 mm.  a previously seen middle lobe nodule is not demonstrated on the current
    study.  a right lower lobe nodule (3:207) now measures 5 x 4 mm, previously 7
    x 3 mm.  no new or enlarging nodules are identified.
    2. airways:  the airways are patent to the subsegmental level.
    3. vessels:  the thoracic aorta and main pulmonary artery are normal in caliber.
    chest cage: no suspicious osseous lesions or acute fractures.
    \\ \hline
    \textbf{Reference:}
    impression:
    1.  no evidence of thoracic metastases.
    2.  please see report from dedicated ct of abdomen and pelvis for
    subdiaphragmatic findings.
    \\ \hline
    \textbf{BART-large:}
    1.  millimetric pulmonary nodules are unchanged since the most recent study, and decreased in size since \_\_\_.
    no new or enlarging pulmonary nodules.
    2.  please see report from dedicated ct of the abdomen and pelvis for subdiaphragmatic findings.
    \\ \textcolor{c_blue1}{(R-1: 57.62; R-2: 45.61; BertScore: 93.27; SummaC: 41.21)}
    \\ \hline
    \textbf{BART-large $+$ L-Prun:}
    1.  millimetric pulmonary nodules are unchanged since the most recent study, and decreased in size since \_\_\_.
    no new or enlarging pulmonary nodules.
    2.  please see report from dedicated ct of the abdomen and pelvis for subdiaphragmatic findings.
    \\ \textcolor{c_blue1}{(R-1: 57.62; R-2: 45.61; BertScore: 93.27; SummaC: 41.21)}
    \\ \hline
    \textbf{BART-large $+$ LoRA$_{FF}$:}
    1. interval decrease in size of several millimetric pulmonary nodules.
    no new or enlarging nodules are identified.
    2. please see report from dedicated ct of the abdomen and pelvis for subdiaphragmatic findings.
    \\ \textcolor{c_blue1}{(R-1: 66.66; R-2: 50.00; BertScore: 92.43; SummaC: 61.83)}
    \\ \hline
    \textbf{BART-large $+$ LoRA$_{FF}$ $+$ L-Prun:}
    1. interval decrease in size of several millimetric pulmonary nodules.
    no new or enlarging nodules identified.
    2. please see report from dedicated ct of the abdomen and pelvis for subdiaphragmatic findings.
    \\ \textcolor{c_blue1}{(R-1: 67.92; R-2: 50.98; BertScore: 93.62; SummaC: 66.12)}
    \\ \hline
    \textbf{T5-large:}
    1. several millimetric pulmonary nodules are unchanged since the most recent study, and decreased in size since the examination from \_\_\_.
    no new or enlarging nodules are identified.
    2. please see report from dedicated ct of the abdomen and pelvis for subdiaphragmatic findings.
    recommendation(s): thyroid nodule. ultrasound follow up recommended.
    \_\_\_ college of radiology guidelines recommend further evaluation for incidental thyroid nodules of 1.0 cm or larger in patients under age \_\_\_ or 1.5 cm in patients age.
    \\ \textcolor{c_blue1}{(R-1: 36.00; R-2: 26.53; BertScore: 88.16; SummaC: 46.79)}
    \\ \hline
    \textbf{T5-large $+$ L-Prun:}
    1. several millimetric pulmonary nodules are unchanged since the most recent examination, and decreased in size since the examination from \_\_\_.
    no new or enlarging nodules are identified.
    2. please see report from dedicated ct of the abdomen and pelvis for subdiaphragmatic findings.
    recommendation(s): thyroid nodule.
    no follow up recommended.
    absent suspicious imaging features, unless there is additional clinical concern, \_\_\_ college of radiology guidelines do not recommend further evaluation for incidental thyroid nodules less than 1.0 cm in
    \\ \textcolor{c_blue1}{(R-1: 35.64; R-2: 26.26; BertScore: 88.33; SummaC: 49.39)}
    \\ \hline
    \textbf{T5-large $+$ LoRA$_{FF}$:}
    1. re-demonstrated are several millimetric pulmonary nodules, unchanged since the most recent study, and decreased in size since the examination from \_\_\_.
    no new or enlarging nodules are identified.
    2. please see report from dedicated ct of the abdomen and pelvis for subdiaphragmatic findings.
    \\ \textcolor{c_blue1}{(R-1: 51.51; R-2: 40.62; BertScore: 90.59; SummaC: 65.19)}
    \\ \hline
    \textbf{T5-large $+$ LoRA$_{FF}$ $+$ L-Prun:}
    1. re-demonstrated are several millimetric pulmonary nodules, unchanged since the most recent examination, and decreased in size since the examination from \_\_\_.
    no new or enlarging nodules are identified.
    2. please see report from dedicated ct of the abdomen and pelvis for subdiaphragmatic findings.
    3. please see report from dedicated ct of the abdomen and pelvis for subdiaphragmatic findings.
    \\ \textcolor{c_blue1}{(R-1: 44.44; R-2: 32.91; BertScore: 89.33; SummaC: 65.74)}
    \\ \hline \\
  \end{tabular}
  \caption{Generated examples of MIMIC-IV-radiology. \# 31.}
  \label{tab:case-rad}
\end{table*}

\begin{table*}[htbp]
  \centering \renewcommand\arraystretch{1.0} \small
  \begin{tabular}{p{0.95\textwidth}}
    \hline
    \textbf{Source:}
    Hello and thank you in advance.  Woke up to a horrible itch on my arm - looked at it -  it is swollen, red about three inches around, very warm to the touch, and now bumpy.  Looks like a spider bite... two tiny red marks in the center.  I cleaned it and sprayed bactine over area.  Is there anything else I can do or be concerned about?
    \\ \hline
    \textbf{Reference:}
    Hi, Welcome to Chat Doctor, It seems that you might have some poisonous insect bite giving allergic rash with severe itching.
    Due to bad itching nail scratches might produce some infection on the part.
    Go for short course of antibiotic to clear infection.
    Take anti-histamine like Cetirizine or Levo-cetrizine as and when required.
    Apply some local triple acting cream.
    Keep that part Chat Doctor.
    Ok and bye.
    \\ \hline
    \textbf{BART-large:}
    Hello, Welcome to Chat Doctor, The history and symptoms are suggesting me that you may be having a fungal infection.
    The fungus is the normal comedians of this area which will be kept under control by the helpful organisms whenever there is an imbalance in the environment this fungus will activate and produces its symptoms.
    I would suggest you to follow.
    1. Keep the area Chat Doctor.
    2. Topical antifungal cream should be applied.
    3. Tab Rifleman, 150 mg single dose will help to control the symptoms.
    Thank you.
    \\ \textcolor{c_blue1}{(R1: 25.47; R-2: 6.45; BLEU-1: 26.00; BLEU-4: 3.09)}
    \\ \hline
    \textbf{BART-large $+$ L-Prun:}
    Hello, Welcome to Chat Doctor, The history and symptoms are suggesting me that you may be having a fungal infection of the area.
    The area is very moist and is prone to develop these infections.
    I would suggest you to follow.
    1. Keep the area clean and Chat Doctor.
    2. Topical antifungal cream should be applied.3.
    Tab Rifleman, 150 mg single dose will help to control the symptoms.
    Thank you.
    \\ \textcolor{c_blue1}{(R1: 28.77; R-2: 7.29; BLEU-1: 31.70; BLEU-4: 3.79)}
    \\ \hline
    \textbf{BART-large $+$ LoRA$_{FF}$:}
    Hi, Welcome to Chat Doctor.
    I have gone through your query and can understand your concerns.
    As per your complaint it seems that you might be having allergic reaction to spider bite.
    I would suggest you to take antihistamine like cetirizine for symptomatic relief.
    You can also apply calamine lotion on the affected area.
    I hope this information has been both informative and helpful for you.
    Wish you good health.
    \\ \textcolor{c_blue1}{(R1: 35.97; R-2: 14.59; BLEU-1: 35.89; BLEU-4: 5.33)}
    \\ \hline
    \textbf{BART-large $+$ LoRA$_{FF}$ $+$ L-Prun:}
    Hello, Welcome to Chat Doctor, I have studied your case.
    As per your symptoms it seems that you might be having fungal infection.
    I would suggest you to consult dermatologist and get examined.
    You may require oral antifungal Chat Doctor.
    You may need oral antihistamines like cetirizine or fluconazole for symptomatic relief.
    I will advise checking your it B12 and it D3 level.
    Hope this answers your query.
    If you have additional questions or follow-up queries then please do not hesitate in writing to us.
    Wishing you good health.
    \\ \textcolor{c_blue1}{(R1: 28.93; R-2: 12.73; BLEU-1: 28.00; BLEU-4: 4.12)}
    \\ \hline
    \textbf{T5-large:}
    Hello, Thank you for posting on Chat Doctor.
    It seems you are suffering from Tina corporal, a kind of fungal infection.
    I would suggest you to consult your dermatologist for proper management of the condition.
    I usually recommend proper course of oral antifungal Chat Doctor.
    You can additionally use antifungal dusting powder containing ketoconazole during daytime and a soap containing ketoconazole for rinsing of affected areas.
    Take oral antihistamines for itching as required.
    Maintain hygiene over those areas and avoid wearing tight undergarments.
    Hope your queries are resolved.
    \\ \textcolor{c_blue1}{(R1: 25.47; R-2: 6.45; BLEU-1: 29.89; BLEU-4: 0.00)}
    \\ \hline
    \textbf{T5-large $+$ L-Prun:}
    Hello, Thank you for posting on Chat Doctor.
    It seems you are suffering from Tina courts and corporal, a kind of fungal infection.
    I would suggest you to consult your dermatologist for proper management of the condition.
    I usually recommend proper course of oral antifungal Chat Doctor.
    You can additionally use antifungal dusting powder containing ketoconazole during daytime and a soap containing ketoconazole for rinsing of affected areas.
    Take oral antihistamines for itching as required.
    Maintain hygiene over those areas and avoid wearing tight undergarments.
    Hope your queries are resolved.
    \\ \textcolor{c_blue1}{(R1: 25.47; R-2: 6.45; BLEU-1: 29.89; BLEU-4: 0.00)}
    \\ \hline
    \textbf{T5-large $+$ LoRA$_{FF}$:}
    Thanks for your question on Chat Doctor.
    I can understand your concern.
    In my opinion, you should consult dermatologist and get done clinical examination of affected area.
    I would suggest you to take antihistamines like ibuprofen, ibuprofen ointment.
    \\ \textcolor{c_blue1}{(R1: 17.09; R-2: 1.73; BLEU-1: 23.33; BLEU-4: 0.00)}
    \\ \hline
    \textbf{T5-large $+$ LoRA$_{FF}$ $+$ L-Prun:}
    Thanks for your question on Chat Doctor.
    I can understand your situation and problem.
    You may need antibiotics and antihistamines.
    You should consult dermatologist and get done clinical examination of affected area.
    You should take treatment accordingly.
    Hope I have answered your question, if you have doubt then I will be happy to answer.
    Thanks for using Chat Doctor.
    Wish you a very good health.
    \\ \textcolor{c_blue1}{(R1: 20.89; R-2: 3.03; BLEU-1: 27.02; BLEU-4: 0.00)}
    \\ \hline \\
  \end{tabular}
  \caption{Generated examples of HealthCareMagic. \# 46.}
  \label{tab:case-hcm}
\end{table*}

\begin{table*}[htbp]
  \centering \renewcommand\arraystretch{1.0} \small
  \begin{tabular}{p{0.95\textwidth}}
    \hline
    \textbf{Source:}
    Hello doctor,I had been diagnosed with hypothyroid nearly 10 months ago. My TSH was over 100 during that time. Also, I had uric acid which was high. Doctor prescribed me to take Thyroxine tablet 50 mg daily and to increase it to 100 mg after one month. Also, uric acid medicine was given. I took that for a couple of months but slowly stopped both medicines. Instead, I am now taking 750 mg Ashwagandha and 400 mg Shilajeet daily.I got my TSH and uric acid tested recently and TSH comes to 12.0 and uric acid is 447 (both results are still higher than normal). Can you guide me for a suitable Ayurveda medicine to bring thyroid and uric acid to a normal level?
    \\ \hline
    \textbf{Reference:}
    Hi.
    I have seen your attachments and read the complaints carefully.
    It is not a big issue and can be managed through Ayurvedic medicines and yoga practices.
    Hormonal imbalance, metabolic disorders.
    TSH (thyroid-stimulating hormone), uric acid, blood pressure, and blood sugar.
    You should take the following medicines regularly for one month and after one month you may revert back with investigation reports.
    Rx, 1. Kanchanar Guggulu one tablet, Gandamala kandan rasa one tablet, Giloy ghanvati one tablet thrice a day after meals with lukewarm water.
    2. Punarnavasava 15 ml with equal amount of water after meal twice a day.
    3. Amla alovera juice 10 ml daily in the morning.
    Avoid spicy and fried junk foods.
    ChatDoctor.
    Avoid milk and milk products, egg and non-vegetarian foods, beans and protien rich food.
    Do regular Kapalbhati and Pranayam like anulom vilom nadi shodhana bhramari and specially ujjai.
    Do these things and provide your valuable feedback after one month.
    \\ \hline
    \textbf{BART-large:}
    Hello. I just read your query and from what you have mentioned it appears that you have hypothyroidism and raised uric acid levels.
    I suggest that you take the following medicines.
    Consult your specialist doctor, discuss with him or her and with their consent take the medicines.1.
    Ashwagandha 750 gm twice a day with warm milk after meals.2.
    Shilajeet 400 mg two times a day.3.
    Avipattikar Guggulu two tablets two times in the morning and two tablets in the evening after food.
    Avoid spicy and fried junk foods.
    \\ \textcolor{c_blue1}{(R1: 37.79; R-2: 12.69; BLEU-1: 54.36; BLEU-4: 8.00)}
    \\ \hline
    \textbf{BART-large $+$ L-Prun:}
    Hello.
    I just read your query and from what you have mentioned it appears that you have hypothyroidism.
    I would recommend that you repeat your thyroid function test after three months and then we will review again.
    I hope this helps you.
    Feel free to write back to me if you have more questions.
    Thanks and regards. For more information consult an ayurveda specialist online
    \\ \textcolor{c_blue1}{(R1: 14.91; R-2: 0.88; BLEU-1: 25.35; BLEU-4: 0.00)}
    \\ \hline
    \textbf{BART-large $+$ LoRA$_{FF}$:}
    Hello. I have gone through your reports.
    Your TSH (thyroid-stimulating hormone) level is high and uric acid level is also high.
    So, you need to take some ayurvedic medicines to bring your thyroid levels back to normal.
    You can take Ashwagandha and Shilajeet.
    It is a good medicine for hypothyroidism.
    But, it is not the best medicine for your problem.
    For more information consult a hematologist online.
    \\ \textcolor{c_blue1}{(R1: 29.53; R-2: 11.06; BLEU-1: 52.32; BLEU-4: 8.43)}
    \\ \hline
    \textbf{BART-large $+$ LoRA$_{FF}$ $+$ L-Prun:}
    Hello. I have gone through your query and understand your concern.
    Your TSH (thyroid-stimulating hormone) level is high and you are taking Ayurveda medicines for hypothyroidism.
    I would like to know a few more details about your thyroid function test.
    If your TSH and uric acid levels are high, then you need to get a thyroid profile done to rule out thyroid disease.
    I suggest you to consult a thyroid specialist for thyroid function tests and get your thyroid profile checked.
    For more information consult a medicine physician online
    \\ \textcolor{c_blue1}{(R1: 21.25; R-2: 3.96; BLEU-1: 32.32; BLEU-4: 2.08)}
    \\ \hline
    \textbf{T5-large:}
    Hello.
    I just read your query and from what you have mentioned it appears that you have hypothyroidism and high uric acid.
    I would like to know a few more details so that I can guide you better.
    Revert back with the answers to the above questions to an ayurveda specialist online.
    \\ \textcolor{c_blue1}{(R1: 18.60; R-2: 2.81; BLEU-1: 36.36; BLEU-4: 0.00)}
    \\ \hline
    \textbf{T5-large $+$ L-Prun:}
    Hello. I have gone through your query and reports (attachment removed to protect patient identity).
    Your TSH (thyroid-stimulating hormone) and uric acid levels are high.
    So, you need to take Ashwagandha and Shilajeet along with Thyroxine.
    \\ \textcolor{c_blue1}{(R1: 27.94; R-2: 8.81; BLEU-1: 53.94; BLEU-4: 9.58)}
    \\ \hline
    \textbf{T5-large $+$ LoRA$_{FF}$:}
    Hello.
    Your TSH (thyroid stimulating hormone) and uric acid levels are high (attachment removed to protect patient identity).
    Thyroxine (Tyrosine sulfate) is an anti-thyroid ChatDoctor.
    It is a steroid hormone.
    It can be taken for a short period of time.
    You can take Ashwagandha and Shilajeet (Ayurveda medicine) for two months and then increase it to 100 mg after one month.
    It will help you to bring your TSH to normal levels.
    For more information consult a gynaecologist online
    \\ \textcolor{c_blue1}{(R1: 29.17; R-2: 11.76; BLEU-1: 43.95; BLEU-4: 6.81)}
    \\ \hline
    \textbf{T5-large $+$ LoRA$_{FF}$ $+$ L-Prun:}
    Hello.
    I have gone through your query and can understand your concern.
    You are suffering from hypothyroidism.
    I would like to know if you have any uric acid problem.
    If you are having hyperthyroiditis, then you need to take Ashwagandha and Shilajeet daily for a few months.
    I suggest you take a tablet of Thyroxine 50 mg daily for one month.
    For more information consult a gynaecologist online.
    \\ \textcolor{c_blue1}{(R1: 21.65; R-2: 4.36; BLEU-1: 41.33; BLEU-4: 0.00)}
    \\ \hline \\
  \end{tabular}
  \caption{Generated examples of iCliniq. \# 184.}
  \label{tab:case-icliniq}
\end{table*}


\end{document}